\documentclass[11pt]{article}

\usepackage{times}

\usepackage{fullpage}
\usepackage[compact]{titlesec}  
\titlespacing{\section}{0pt}{0pt}{0pt}
\titleformat{\section}
  {\normalfont\fontsize{13}{15}\bfseries}{\thesection}{1em}{}
\titleformat{\section}
  {\normalfont\fontsize{13}{15}\bfseries}{\thesection}{1em}{}

\usepackage[utf8]{inputenc}
\usepackage{booktabs}
\usepackage{enumitem}
\usepackage{authblk}

\title{Using DeepSpeed and Megatron to Train Megatron-Turing NLG 530B, A Large-Scale Generative Language Model}

\author[$\mathsection$,$\dag$]{Shaden Smith}
\author[$\mathsection$,$\ddag$]{Mostofa Patwary}
\author[$\dag$]{Brandon Norick}
\author[$\ddag$]{Patrick LeGresley}
\author[$\dag$]{Samyam Rajbhandari}
\author[$\ddag$]{Jared Casper}
\author[$\dag$]{Zhun Liu}
\author[$\ddag$]{Shrimai Prabhumoye}
\author[$\dag$]{George Zerveas\footnote{Affiliated with Brown University. Work done during internship at Microsoft.}}
\author[$\ddag$]{Vijay Korthikanti}
\author[$\dag$]{Elton Zhang}
\author[$\ddag$]{Rewon Child}
\author[$\dag$]{Reza Yazdani Aminabadi}
\author[$\ddag$]{Julie Bernauer}
\author[$\dag$]{Xia Song}
\author[$\ddag$]{Mohammad Shoeybi}
\author[$\dag$]{Yuxiong He}
\author[$\ddag$]{Michael Houston}
\author[$\dag$]{Saurabh Tiwary}
\author[$\ddag$]{Bryan Catanzaro}
\affil[$\mathsection$]{equal contribution}
\affil[$\dag$]{Microsoft}
\affil[$\ddag$]{NVIDIA}

\date{}

\usepackage{adjustbox}
\usepackage{tabularx}
\usepackage{todonotes}
\usepackage{footnote}
\usepackage[numbers]{natbib}
\usepackage{graphicx}
\usepackage{hyperref}
\usepackage{listings}
\usepackage{placeins}
\usepackage{amsmath}

\definecolor{dkgreen}{rgb}{0,0.6,0}
\definecolor{gray}{rgb}{0.5,0.5,0.5}
\definecolor{mauve}{rgb}{0.58,0,0.82}

\newcommand{\ours}{MT-NLG}

\graphicspath{{images/}}

\begin{document}

\parindent 0.0in
\parskip 0.15in

\maketitle

\begin{abstract}

Pretrained general-purpose language models can achieve state-of-the-art accuracies in various natural language processing domains by adapting to downstream tasks via zero-shot, few-shot and fine-tuning techniques. Because of their success, the size of these models has increased rapidly, requiring high-performance hardware, software, and algorithmic techniques to enable training such large models. As the result of a joint effort between Microsoft and NVIDIA, we present details on the training of the largest monolithic transformer based language model, Megatron-Turing NLG 530B ({\ours}), with 530 billion parameters. In this paper, we first focus on the infrastructure as well as the 3D parallelism methodology used to train this model using DeepSpeed and Megatron. Next, we detail the training process, the design of our training corpus, and our data curation techniques, which we believe is a key ingredient to the success of the model. Finally, we discuss various evaluation results, as well as other interesting observations and new properties exhibited by {\ours}. We demonstrate that {\ours} achieves superior zero-, one-, and  few-shot learning  accuracies on  several  NLP  benchmarks and establishes new state-of-the-art results. We believe that our contributions will help further the development of large-scale training infrastructures, large-scale language models, and natural language generations.
\end{abstract}

\section{Introduction}

The recently released \emph{foundation models}~\cite{bommasani2021opportunities}, such as BERT \cite{Devlin2019BERTPO}, GPT-2~\cite{gpt2-radford2019language}, and RoBERTa~\cite{liu2019roberta}, represent a paradigm shift in which AI systems can be built by pretraining a general class of models at scale and then adapting them for a wide range of downstream tasks through transfer learning. Such models became ubiquitous in state-of-the-art natural language processing (NLP) systems by embracing the effectiveness of a combination of factors: the transformer architecture~\cite{DBLP:journals/corr/VaswaniSPUJGKP17}, self-supervised learning, few-shot conditioning~\cite{brown2020language}, and fine-tuning.


Importantly, many recent works have established that scaling up models greatly improves their performance, with especially substantial performance improvements in the zero-shot and few-shot settings. For example, GPT-3 \cite{brown2020language}, an autoregressive language model with 175~billion parameters, performs competitively on language tasks using in-context learning without fine-tuning or gradient updates. Such in-context learning allows models to perform new language tasks with only simple instructions and a few optional examples. The effectiveness of this method was further enhanced by recent model adaptation work such as prompt tuning \cite{lester2021power}, which efficiently adapts large language models to individual tasks with robust task performance. Other intriguing capabilities exhibited by large language models include, but are not limited to, free-form generation of coherent, long-form text like news stories, generating responses with real-world knowledge, as well as performing rudimentary mathematical operations.

\begin{figure}
    \centering
    \includegraphics[width=0.65\linewidth]{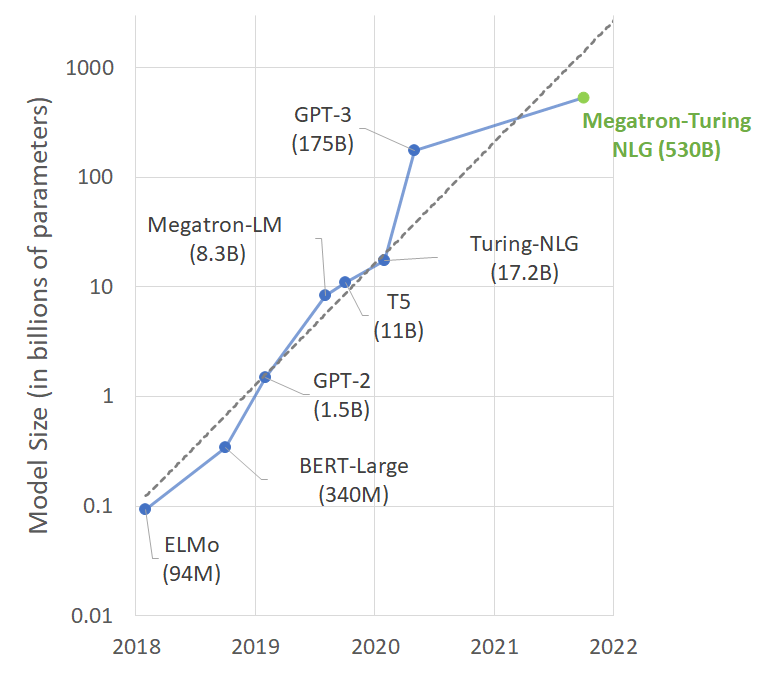}
    \caption{Trend of sizes of state-of-the-art NLP models with time.}
    \label{fig:nlp_trends}
\end{figure}

The rapid development of large language models in recent years has also been fueled by growth in computational resources, availability of large datasets and evolving software stacks. State-of-the-art supercomputing clusters address the computation, memory and networking need of model training at this scale. Careful processing of high-quality, high-volume and diverse datasets directly contributes to model performance in downstream tasks as well as model convergence. New approaches to numerical manipulation and training recipes were developed aiming at improved optimization efficiency and stability. However, to sustain the seemingly exponential growth of model parameter size (see Figure \ref{fig:nlp_trends}), substantial progress in developing new methods, infrastructure and training capabilities is needed.

Training such large models is challenging for two reasons.  First, it is no longer possible to fit the parameters of these models in the memory of even the largest GPU.
Second, the large number of compute operations required can result in unrealistically long training times if special attention is not paid to concurrently optimizing the algorithms, software, and hardware stack. This calls for efficient parallelism techniques scalable on both memory and compute, in order to achieve the full potential of thousands of GPUs.

Compelled by the impressive qualitative performance improvements owing to an increasing model size that have been previously exhibited, our work continues the trend of large-scale language modeling. We built Megatron-Turing NLG 530B ({\ours}), a transformer-based language model with 530~billion  parameters. It is, to the best of our knowledge, the largest monolithic language model trained to date, with 3x more parameters than GPT-3. It is worth noting that sparse models structures encompassing a higher total number of parameters, such as mixture-of-experts \cite{Shazeer2017OutrageouslyLN}, have been trained. However, it is unclear whether models built following this approach would have comparable parameter efficiency and generalization capability.

Training {\ours} was made feasible by numerous innovations and breakthroughs along all AI axes. Through a collaboration between NVIDIA Megatron-LM \cite{megatron-DBLP:journals/corr/abs-1909-08053, Megatron_2021} and Microsoft DeepSpeed~\cite{rasley2020deepspeed,deepspeedgithub}, we created an efficient and scalable 3D parallel system capable of combining data, pipeline, and tensor-slicing based parallelism. By combining tensor-slicing and pipeline parallelism, we can operate within the regime where they are most effective. We built high-quality, natural language training corpora with hundreds of billions of tokens, and co-developed training recipes to improve optimization efficiency and stability.

In this paper, we will discuss details of our methods during the development of {\ours}, including training infrastructure (Section \ref{sec:train_infra}), training dataset and training process (Section \ref{sec:datamodel}), model evaluation and other interesting observations (Section \ref{sec:results}). We will also present an in-depth study on social biases (Section \ref{sec:biases}), in-context learning capability (Section \ref{sec:hans}) and qualitative analysis of the generation capability (Section \ref{sec:gen_capa}) of {\ours}. 



\section{Large Model Training Infrastructure}
\label{sec:train_infra}

Powered by NVIDIA A100 Tensor Core GPUs and HDR InfiniBand networking, state-of-art clusters (such as NVIDIA Selene and Microsoft Azure NDv4) have enough compute power to train models with trillions of parameters. However, achieving the full potential of these supercomputers requires memory- and compute-efficient strategies for parallelizing across thousands of GPUs. 
In isolation, existing parallelism strategies such as data, pipeline, or tensor-slicing have trade-offs in memory and compute efficiency and cannot be used to train models at this scale. In this section, we discuss the system challenges of training large models.  We describe our software design, hardware system, and the performance evaluation of a unified, powerful training infrastructure.  

\subsection{Challenges}

We begin by discussing the challenges of training large-scale language models: memory and compute efficiency, and the tradeoffs of various solution strategies such as data, tensor and pipeline parallelism.

\subsubsection{Memory and Compute Efficiency}
    
\paragraph{Memory Efficiency} The memory requirements to train a 530~billion parameter
model are far beyond what is available on a single GPU device. 
We refer to Rajbhandari~et~al.~\cite{rajbhandari2021zero} for an analytical
study of memory consumption during training.

Mixed precision training~\cite{micikevicius2017mixed} typically stores weights and gradients in half precision formats (i.e., 2~bytes per parameter) 
for forward and backward propagations. It also keeps full-precision (4~bytes) copies in 32 bit float format for numerical stability
in the optimizer. Assuming training with the Adam optimizer~\cite{kingma2014adam},
training consumess 20 bytes of memory per parameter:
$$
    \underbrace{2 + 4}_{\text{weights}} 
        + \underbrace{2 + 4}_{\text{gradients}}
        + \underbrace{4 + 4}_{\text{Adam states}}
    = 20\ \text{bytes}.
$$
Training a 530~billion parameter model thus requires over 10~terabytes of aggregate memory for the model weights, gradients, and optimizer states.

Activations can also consume significant memory and scale with the training batch size, sequence length,
and model dimensions. Checkpointing and recomputing the activations of
each transformer block is a common strategy for training large language models to reduce the memory required for activations.
However, the activations at the boundary between layers still needs to be stored and the aggregate activation memory is:
$$
    \text{batch-size} \times \text{number-of-layers} \times \text{sequence-length} \times \text{hidden-dimension} \times 2\ \text{bytes},
$$
which is approximately 16.9~terabytes following our model and training configuration (Section~\ref{sec:modelconfig}).

Fortunately, activation memory requirements can be mitigated by virtue of
\emph{gradient accumulation}. Gradient accumulation is a strategy in which the full training
batch is split into micro-batches that are processed in sequence and their resulting gradients are
accumulated before updating the model weights. After computing the gradient for a micro-batch, the
associated activations can be freed. As a result, the training batch size can scale without
increasing the peak resident activation memory. For example, training with 1920 micro-batches
instead of a single micro-batch of size 1920 reduces the peak activation memory from 16.9~terabytes
to 8.8~gigabytes without changing the effective batch size. 

\paragraph{Compute Efficiency} While large GPU clusters can have thousands of high-throughput
GPUs, achieving high compute efficiency at this scale is challenging. A large batch
size can be an effective way of increasing compute efficiency, because it increases the arithmetic
intensity of a kernel and helps amortize the time spent stalled on communication and synchronization.
However, the batch size that a model can be
trained with has an upper bound; using too large of a batch size can have
negative effects on the model quality. With 4000 GPUs, even a large batch
size of 4000 would only allow for a batch size of 1 per GPU and limit
compute efficiency.

\subsubsection{Tradeoffs of Data, Tensor, and Pipeline Parallelism}

\paragraph{Data Parallelism}
Data parallelism is a ubiquitous technique in deep learning in which
each input batch of training data is divided among the data-parallel workers.
Gradients are communicated and aggregated among data-parallel workers before
updating the model weights.
Data parallelism has
several distinct advantages, including compute efficiency and ease of implementation. However, data parallelism relies on scaling the batch
size with the number of data-parallel workers, and cannot be made arbitrarily
large without affecting model quality.

\textit{Memory Efficiency}: Data parallelism replicates the model and optimizer
across all workers, and therefore is not memory efficient.
The Zero Redundancy Optimizer (ZeRO)~\cite{rajbhandari2020zero} is a collection
of optimizations that improve the memory efficiency of data parallelism by 
partitioning the replicated data among data-parallel workers.

\textit{Compute Efficiency}: The amount of computation performed by each worker
is constant as we increase the degree of parallelism and training batch size.
Data parallelism can achieve near-perfect scaling at small scales.
However, the communication cost of
aggregating gradients increases with the model size and
can limit compute efficiency on large models or systems with low communication
bandwidth.
Gradient accumulation is also a common strategy for amortizing this
communication cost by further increasing the batch size and performing multiple
forward and backward propagations on micro-batches while locally accumulating
gradients before aggregating and taking an optimizer step. Additionally,
performance can be increased by simultaneously communicating gradients that have
already been communicated in parallel with computing the gradients for other tensors.

\paragraph{Tensor Model Parallelism}
Tensor model parallelism (or, \emph{tensor parallelism})
is a broad class of model parallelism techniques that partitions the
individual layers of the model across workers. Tensor parallelism reduces the
memory proportional to the number of workers. Megatron~\cite{megatron-DBLP:journals/corr/abs-1909-08053} uses model parallelism
to efficiently partition transformer blocks for large-scale language models.

\textit{Memory Efficiency}: Tensor parallelism reduces the memory footprint of
the model proportional to the number of workers. Depending on the model
architecture, some of the activation memory is also reduced, although there may
still be some replications.

\textit{Compute Efficiency}: Tensor parallelism introduces additional communication of activations in each forward and backward
propagation. Therefore, tensor parallelism requires high communication bandwidth to be
efficient and is best kept within a single DGX sever where high bandwidth NVLink is available. Furthermore, each model-parallel worker decreases the
amount of computation performed between each communication stage, impacting
compute efficiency. Tensor parallelism is often used to expand the envelope of memory and compute efficiency beyond what data parallelism alone can do.

\paragraph{Pipeline Model Parallelism}
Pipeline model parallelism (or, \emph{pipeline parallelism}) divides the layers of the model into stages that
can be processed in parallel~\cite{huang2019gpipe,narayanan2019pipedream}.
As one
stage completes the forward pass for a
micro-batch, the activation memory is communicated to the next stage in the
pipeline. Similarly, as the next stage completes its backward propagation,
gradients are communicated backwards through the pipeline. Multiple
micro-batches must be kept in flight to ensure pipeline stages compute in
parallel.

\textit{Memory Efficiency}: Pipeline parallelism reduces memory proportionally to
the number of pipeline stages, allowing model size to scale linearly with the
number of workers. However, pipeline parallelism does not reduce the memory
footprint for the activations of each layer. Additionally, each worker must
store the activations for all micro-batches in flight. We use a 1F1B pipeline
schedule~\cite{narayanan2019pipedream} that alternates forward and backward
propagations. A key benefit of 1F1B is that the number of micro-batches in flight
is bounded by the number of pipeline stages, as opposed to the total number of micro-batches
in a full training batch. 

\textit{Compute Efficiency}: Pipeline parallelism has the smallest communication
overhead of the three approaches, as it only communicates the activations 
between the pipeline stage boundaries.
However, it cannot scale indefinitely. The degree of pipeline parallelism is 
bounded by the depth of the model, and increasing the pipeline dimension
decreases the compute efficiency like other forms of model parallelism.
Pipeline parallelism also requires each of its stages to be load
balanced for high efficiency.

Pipeline parallelism incurs a bubble overhead from filling and
emptying the pipeline at the beginning and end of each training batch. The size
of the bubble overhead bounds the potential speedup from pipeline parallelism.
The fraction of perfect speedup achievable (or, \emph{parallel efficiency}) 
is a function of the number of pipeline stages ($PP$) and total micro-batches ($MB$):
$$
    \text{efficiency} = \frac{MB}{MB + PP - 1}.
$$
If the number of micro-batches is 4x or 8x the number of pipeline stages,
the pipeline achieves 81\% and 90\% parallel efficiency from one  pipeline stage,
respectively.

From the above discussion, it is clear that none of the existing parallelism
techniques can address all the system challenges of training models with
hundreds of billions of parameters. However, each
parallelism technique has its own merits and can be used in a complementary
fashion.
To this end, we use \emph{3D parallelism}, which is a systematic combination of data, tensor, and pipeline
parallelism that addresses both compute and memory efficiency
simultaneously.

\subsection{Software System --- 3D Parallelism with DeepSpeed and Megatron }

Our system software stack combines pipeline parallelism and data parallelism
from DeepSpeed with tensor-slicing from Megatron to create a flexible
3D-parallelism implementation. 
Data, tensor, and pipeline
parallelism each play a specific role in improving memory and compute
efficiency.

\textit{Memory Efficiency}: Transformer blocks are divided into pipeline
stages, and the blocks of each stage are further divided via tensor parallelism.
This 2D combination simultaneously reduces the memory consumed by the weights,
gradients, optimizer states, and activations. However, we cannot partition the model indefinitely without losing compute efficiency.

\textit{Compute Efficiency}: To further accelerate training,
we use data parallelism to scale to arbitrarily large number of GPUs.
For example, each 530~billion parameter model replica spans 280 NVIDIA A100 GPUs, with 8-way
tensor-slicing within a node and 35-way pipeline parallelism across nodes.
We then use data parallelism to scale out further to thousands of GPUs.

Our 3D parallelism implementation is optimized using topology aware mapping,
which minimizes communication overhead across all forms of parallelism, and has an  especially large impact on data parallelism. This mapping is key to achieving excellent
compute efficiency at scale. We discuss the details below.

\subsubsection{Topology-Aware 3D Mapping} 

Each axis of parallelism is carefully mapped onto the workers to maximize
compute efficiency by exploiting two key architectural properties.

\paragraph{Mapping for Bandwidth}
Intra-node communication has a higher bandwidth than inter-node.
We prioritize co-locating parallel groups with larger communication volumes
in order to utilize the higher bandwidth.
Tensor parallelism has the largest communication overhead of the three strategies,
and so we prioritize placing tensor parallel workers within a node.
When possible, data parallel workers are also placed
within a node to accelerate gradient communications. Otherwise, data parallel workers
are mapped to nearby nodes when possible.
Pipeline parallelism has the lowest communication volume, and so we can schedule pipeline
stages across nodes without being limited by the communication bandwidth.

\paragraph{Bandwidth Amplification}
The volume of gradient communication by each data parallel group decreases linearly
as pipeline and tensor parallelism increase. Thus, the total communication volume is
decreased from pure data parallelism. Furthermore, each data parallel group
performs its communication independently and in parallel among a subset of more
localized workers. As a result, the effective bandwidth for data parallel
communication is amplified by a combination of reduced communication volume and
increased locality and parallelism.

\subsection{Hardware System}
Model training is done with mixed precision using 16-bit bfloat on NVIDIA’s Selene \cite{selene}
supercomputer with 560 DGX A100 nodes. Each cluster node has 8 NVIDIA 80-GB A100
GPUs \cite{a100}, connected to each other by NVLink and NVSwitch \cite{nvlink}.
Each node has eight NVIDIA Mellanox 200Gbps HDR Infiniband HCAs for application
communication, with an additional two HCAs per node for dedicated storage. The
nodes are connected in a three-level (leaf, spine, core) fat-tree topology with
850 switches. This topology allows efficient all-reduce communication (which is the dominant
communication pattern in deep learning training). The cluster uses an all-NVME
shared parallel filesystem for high-performance data access and storage. The
peak device throughput of an A100 GPU with 16-bit precision is 312 teraFLOP/s,
resulting in an aggregate of 1.4 exaFLOP/s of peak 16-bit precision performance.

\subsection{System Performance Evaluation}
We considered the end-to-end throughput of our system for the 530~billion parameter model with batch size 1920 on 280, 350, and 420 DGX A100 servers on Selene. We observed iteration times of 60.1, 50.2, and 44.4 seconds, respectively. These correspond to 126, 121, and 113 teraFLOP/s per GPU, respectively.

\section{Training Dataset and Model Configuration}
\label{sec:datamodel}
In this section we present details on the training datasets, our preprocessing techniques, and the model and hyperparameters used in our experiments.

\subsection{Training Dataset and Preprocessing}
Resources such as Common Crawl (CC) provide snapshots of the web which can be utilized as a source of language data. While these data sources contain an enormous amount of language data, they also require carefully designed preprocessing steps in order to select data which is of reasonable quality. As prior work has found (e.g., \citep{brown2020language}), the quality of unfiltered Common Crawl data is lower than that of curated datasets and steps should be taken to increase the average quality of  data selected from Common Crawl for LM pretraining. In addition to CC data, there are many other high quality data sources on the web. To compile our training dataset, we made use of recent work aimed at collecting a diverse training set for language modeling \citep{pile-dataset-2020}. We additionally included RealNews \cite{DBLP:journals/corr/abs-1905-12616} and CC-Stories~\cite{trinh18} which have previously been used for large LM pretraining \cite{tnlg17b,megatron-DBLP:journals/corr/abs-1909-08053}.

\subsubsection{Training Dataset}
We largely built upon prior work described in \citep{brown2020language,pile-dataset-2020} to generate our training set. First, we selected a subset of the datasets from The Pile that we observed to be of the highest relative quality (see Table \ref{tab:datasets-530b}). Then, following a similar approach as that used to generate Pile-CC in \citep{pile-dataset-2020}, we downloaded and filtered two full CC snapshots (2020-50 and 2021-04). At a high level, the steps taken for CC data include text extraction from raw HTML provided in WARC files, scoring extracted documents using a classifier trained on high quality data, and filtering documents according to their scores. These steps are covered in more detail in Section~\ref{preprocessing-details}. Finally, we used fuzzy deduplication to remove duplicate and near duplicate documents from the entire dataset as well as $n$-gram based filtering to remove downstream task data in order to avoid contamination.

\begin{table}[t]
\centering
\small
\begin{tabular}{l c c c}
 \toprule
 \textbf{Dataset} &  \textbf{Tokens (billion)}  &  \textbf{Weights (\%)} &  \textbf{Epochs} \\
 \midrule
 Books3	&	25.7	&	14.3	&	1.5	\\
 OpenWebText2	&	14.8	&	19.3	&	3.6	\\
 Stack Exchange	&	11.6	&	5.7	&	1.4	\\
 PubMed Abstracts	&	4.4	&	2.9	&	1.8	\\
 Wikipedia	&	4.2	&	4.8	&	3.2	\\
 Gutenberg (PG-19)	&	2.7	&	0.9	&	0.9	\\
 BookCorpus2	&	1.5	&	1.0	&	1.8	\\
 NIH ExPorter	&	0.3	&	0.2	&	1.8	\\
 ArXiv	&	20.8	&	1.4	&	0.2	\\
 GitHub	&	24.3	&	1.6	&	0.2 \\
 Pile-CC	&	49.8	&	9.4	&	0.5	\\
 \midrule
 CC-2020-50	&	68.7	&	13.0	&	0.5	\\
 CC-2021-04	&	82.6	&	15.7	&	0.5	\\
 \midrule
 Realnews	&	21.9	&	9.0	&	1.1	\\
 CC-Stories	&	5.3	&	0.9	&	0.5	\\
 \bottomrule
\end{tabular}
\caption{Datasets used to train the MT-NLG model. The top 11 rows are from the Pile dataset, followed by two Common Crawl snapshots, RealNews, and CC-Stories datasets.}
\label{tab:datasets-530b}
\end{table}

\subsubsection{Pre-Processing Details}
\label{preprocessing-details}
\textbf{Common Crawl}: As mentioned previously, Common Crawl comprises an immense amount of data. We chose to process two snapshots, \textit{2020-50} and \textit{2021-04}, with the aim of acquiring around 150B tokens of training data. The first step of this process is language detection \cite{danilk-2021-langdetect} and text extraction from the raw HTML included in the Common Crawl WARC files\footnote{\url{https://github.com/leogao2/commoncrawl\_downloader}}. Following the rationale presented in \citep{danilk-2021-langdetect}, we used the pycld2\footnote{\url{https://pypi.org/project/pycld2/}} and jusText\footnote{\url{https://pypi.org/project/jusText/}} 
libraries for these tasks. We observe that the language detection and extraction step reduces the number of documents significantly, with only around 25\% of documents being classified as English and having non-empty body content.

In order to select high quality documents from these extractions, we trained a 2-gram fastText \citep{ortizsuarez:hal-02148693} classifier. For positive documents, we randomly select 500000, 295000, and 5000 documents from OpenWebText2, Wikipedia, and Books3, respectively, similar to \cite{brown2020language}. 
For negative documents, we randomly sampled an equal number of documents from the text extraction output described above. We held out 10\% of these documents for evaluation of the classifier, which achieved an accuracy of 90.3\% on the held out set after training. The classifier was applied to each of the extracted documents and the probability of the positive label was taken as the score for the document.

Using the scores produced by the process above, we filtered the extracted documents with a Pareto distribution with $\alpha=3$. This resulted in around 80\% of text content being filtered. While our choice of $\alpha$ is lower than some previous works \citep{brown2020language}, manual inspection of the data indicated that it was of acceptable quality and the use of $\alpha = 3$ allowed us to reach and slightly exceed our original token goal after deduplication.




\textbf{Other Datasets}: In addition to Common Crawl data, we leveraged a number of other previously generated datasets. From The Pile, we selected Books3, OpenWebText2, Stack Exchange, PubMed Abstracts, Wikipedia, Gutenberg (PG-19), BookCorpus2, NIH ExPorter, and Pile-CC datasets. We also included the CC-Stories and RealNews datasets used to train Megatron \citep{megatron-DBLP:journals/corr/abs-1909-08053}. For detailed discussions of the preprocessing used for these datasets, we refer to \cite{pile-dataset-2020}.

\textbf{Fuzzy Document Deduplication}: Content on the internet is often duplicated across many documents. To compound this issue, the URLs scraped in different Common Crawl snapshots are not necessarily unique. Indeed, for the snapshots we chose 53\% and 34\% of documents come from new URLs not seen in previous snapshots. Furthermore, it is likely that content contained in our other datasets, such as web content from OpenWebText2 or Wikipedia, will also exist in Commom Crawl.

Exact match duplicates would be computationally expensive,
so we opted to take a fuzzy deduplication approach similar to other works \citep{brown2020language,pile-dataset-2020}. We used a hashing vectorizer with 1,048,576 features to vectorize documents (HashingVectorizer from \texttt{scikit-learn}\footnote{\url{https://scikit-learn.org/stable/}}), calculated min-hashes of the vectorized documents (using \texttt{datasketch}\footnote{\url{http://ekzhu.com/datasketch/documentation.html}}), and performed Locality Sensitive Hashing (LSH) through \texttt{datasketch} on all min-hashes in order to identify potential duplicates. We set our LSH parameters in such a way as to increase the likelihood that documents with Jaccard similarity $\ge$ 0.8 would occur in at least one LSH bucket together. Specifically, we used 20 bands of size 13 for a total of 260 hash functions.

After performing LSH, we processed each bucket and calculated an approximation of the all-pairs Jaccard similarity in order to remove false positive duplicates introduced by LSH. This approximation consisted of $i=0..10$ iterations of sampling a random document $d_i$, calculating the Jaccard similarity with everything remaining in the bucket, removing those documents above the 0.8 threshold and marking them as duplicates of $d_i$. After all buckets were processed and duplicates (at the threshold) were approximately discovered, we constructed a sparse document graph and found the connected components therein (using \texttt{scipy}). Each connected component represents a set of documents that we consider similar enough to be duplicates, and from which we select a single representative. Because the datasets are of varying quality, we defined a priority order based on which dataset to use when selecting representative documents, and the first document encountered from the highest priority dataset within each component was ultimately kept, while the rest were discarded.

\textbf{Additional Processing}: We use the \texttt{Ftfy} library \cite{speer-2019-ftfy} on the training dataset to convert bad unicode text to good unicode text. Additionally, we use the \texttt{langdetect} \cite{danilk-2021-langdetect} library to identify non-English documents and remove any document such with less than 512 characters. If a training document contains the word ``javascript'' and has less than 256 characters, we remove that document as well. 

\textbf{Downstream Task Data Removal}: We use $n$-grams to remove texts that occur in the downstream tasks from the training datasets. When we find an $n$-gram match between a task document and a training document, we split the training document into two pieces by removing the $n$-gram along with 200 characters from both of its sides. We also remove any split training document with fewer than 200 characters, or training documents which were split more than 10 times. Our deduplication process and the values of $n$ used for different tasks are similar to \cite{brown2020language}. Out of 319,781,622 documents from the 15 deduplicated datasets mentioned above, during task deduplication 35,988 documents were split, 1,109 documents were removed, 54 documents were split more than 10 times, and 9,891 were trimmed at the beginning or the end.

\textbf{Blending Datasets}: We opted to blend the datasets into heterogeneous batches according to the sampling weights given in Table~\ref{tab:datasets-530b}. However, the mixing weights do not result in an even split of the samples in each batch for our chosen batch size. To resolve this issue, we track the under- and oversampling for each dataset and slightly adjust the batch composition at each step in order to maintain a sample distribution as close as possible to the chosen mixing weight distribution. 

\subsection{Model and Training Process}
\label{sec:modelconfig}

We used the architecture of the transformer decoder \cite{gpt2-radford2019language}, which is a left-to-right, autoregressive, generative transformer-based language model, and scaled it up to 530~billion parameters.
The number of layers, hidden dimensions, attention heads are 105, 20480, and 128, respectively. The sequence length is 2048 and the global batch size is 1920. We used 8-way tensor and 35-way pipeline parallelism. The learning rate is $5.0 e^{-5}$. We used one billion tokens for linear learning rate warmup. We used cosine decay for the learning rate targeting to reach 10\% of its value over 340~billion tokens. Over the first 12~billion tokens, we started at a batch size of 32 and gradually increased the batch size in increments of 32, until we reach the final batch size of 1920.
We used Adam optimizer with $\beta_1 = 0.9$, $\beta_2 = 0.95$, and $\epsilon = 10^{-8}$. We clipped the gradient norm at 1.0 and used a weight decay of 0.1. For weight initialization, we used a normal distribution with zero mean and a standard deviation of $4.0 e^{-3}$. Our training dataset consists of 339 billion tokens and we trained {\ours} on 270 billions tokens  by blending the 15 training datasets as described above. We also set aside 2\% of our data for validation.

At the scale of models such as {\ours}, training stability is a fundamental challenge. While training the model, we observed that the learning rate, weight initialization, and Adam optimizer parameters directly affect model stability. We projected the learning rate for {\ours} by plotting the learning rates with the size of the models in \cite{brown2020language}. Higher 
learning rate increases the model instability. We used approximately $\sqrt{1/(3*H)}$ as a standard deviation for weight initialization, where $H$ denotes the size of the hidden dimension. Similar to \cite{DBLP:journals/corr/abs-1910-05895}, we also observed that using higher variance for weight initialization fails to converge. We also reduced $\beta_2$ from its standard value of $0.99$ to reduce spikes in the training loss. 


\section{Results and Achievements}
\label{sec:results}

\begin{figure}
    \centering
    \includegraphics[width=0.8\linewidth]{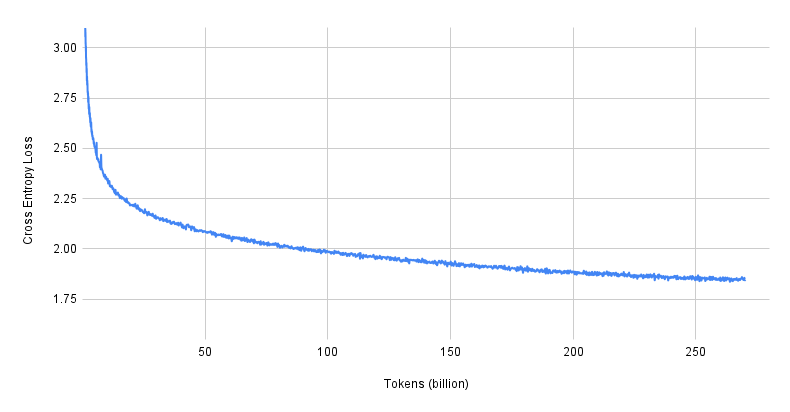}
    \caption{Validation loss of {\ours}.}
    \label{fig:loss_curve}
\end{figure}

To provide a better understanding of how language model performance improves during training, we first present the validation loss curve (cross entropy) of {\ours} in Figure \ref{fig:loss_curve}. Our validation dataset consists of 5.5 billion tokens, so measuring the loss using the entire dataset is computationally expensive. We therefore shuffle the sequences in the validation dataset and then during each computation of validation loss, we run four iterations with global batch size of 1920. This leads to evaluating on a total of 16 million consecutive tokens for each loss computation.

The validation cross-entropy loss is 
3.15 after the model is trained on the first 1 billion tokens. As mentioned earlier, we increase the batch size linearly over the first 12~billion tokens. At the end of this phase, the loss becomes 2.31. When the model reaches our targeted number of tokens, 270~billion, the validation loss becomes 1.85. 

To evaluate the quality of our model (as well as other pretrained language models), we adopt a zero-/one-/few-shot evaluation setting similar to prior work~\citep{brown2020language,Rae2021Gopher}. For better reproducibility, we base our evaluation on the open-source project, {\texttt {lm-evaluation-harness}}~\citep{eval-harness}, and made task-specific changes as appropriate to align our setting more closely with prior work. We will discuss any idiosyncrasies of each task in the task-specific paragraphs. In addition, for our few-shot experiments, we do not do any search for the optimal number of shots, and directly use the configurations suggested in \cite{brown2020language}. In most cases, they seem to perform sufficiently well.

To ensure the evaluation is comprehensive, we choose eight tasks from five different categories: completion prediction, reading comprehension, commonsense reasoning, natural language inference and word sense disambiguation. We present comparisons on these tasks with previous works on pretrained large language models, while also providing supervised baselines whenever applicable to provide context for the gap between ``generalist'' models like pretrained language models and ``specialist'' models that are finetuned on the target task.

Many evaluation tasks involve scoring candidate completion sentences with the model. Unless otherwise stated, the ``likelihood" mentioned in the following context refers to the probability of the candidate answer (conditioned on the prompt) normalized by its number of tokens.

\subsection{Completion Prediction}

\paragraph{LAMBADA} The LAMBADA~\citep{paperno-etal-2016-lambada} dataset is a collection of narrative passages, specifically selected such that a human can easily guess the last word if the whole passage is given as context, but would not be able to answer if only given the last sentence in the passage. This task tests language models' capabilities to understand and retain information from a broader discourse context, instead of just relying on local context or simple statistical patterns.

When evaluating this task zero-shot, we feed each passage to the model as input and check if the model can produce the correct last word via greedy generation (picking tokens with maximum probability). However, for one-/few-shot evaluations, we switched over to a cloze-style prompt format to better suggest to the model that the task is about predicting the \emph{last word} of a sentence as opposed to arbitrary plausible continuation. In such a case, we would insert ``\texttt{\_\_\_\_. $\rightarrow$ }'' before the last word, e.g. ``\texttt{... Paul and Debbie looked at each other, then at \_\_\_\_. $\rightarrow$ Bob}'' and examine if the model would predict the correct word after the ``$\rightarrow$''. We observe significant performance boost in few-shot settings with the cloze-style prompting, although one-shot performance takes a hit, which aligns with observations from prior work~\citep{brown2020language}. Our model's performance in terms of accuracy is shown in table \ref{tab:lambada-perf}, and we are establishing new state-of-the-arts on LAMBADA for all 3 settings on its test set.


\begin{table}[t]
\centering
\small
\begin{tabular}{l c c c}
 \toprule
 & \multicolumn{3}{c}{LAMBADA (acc)} \\
 \cmidrule{2-4}
 \textbf{Model} &  \textbf{Zero-shot}  &  \textbf{One-shot} &  \textbf{Few-shot} \\
 \midrule
 GPT-3  &   76.20   &   72.50   &   86.40  \\
 Gopher  &   74.50   &   -   &   -  \\
 \ours~(ours)	&	\textbf{76.56}	&	\textbf{73.06}	&	\textbf{87.15}	\\
 \bottomrule
\end{tabular}
\caption{LAMBADA zero-shot, one-shot and few-shot accuracy. \ours~outperforms previous models across different settings and establishes new SOTA for all 3 settings. We did not find any recent strong supervised baseline for LAMBADA, hence we omit the comparison with supervised models here.}
\label{tab:lambada-perf}
\end{table}

\subsection{Reading Comprehension}

In this section, we discuss the evaluation of \ours~for reading comprehension. We selected two datasets targeting different styles of questions, and have found very different trends when we increase the number of examples for them during evaluation.

\paragraph{RACE} RACE~\citep{lai-etal-2017-race} is a large-scale reading comprehension dataset, whose passages and questions are extracted from English examinations. Each example in this task consists of an article and several question-answer pairs. To construct prompts, we prepend ``\texttt{Article: }'', ``\texttt{Question: }'', and ``\texttt{Answer: }'' tags to the article, questions and answers text respectively and join them together with a newline in between. The actual answer to the last question is removed, ending the prompt at the last ``\texttt{Answer:}''. We then use the model to score all possible candidate answers as continuations after ``\texttt{Answer:}'' and pick the highest-scoring one as the model's choice.

There are two question types in this dataset: direct questions (e.g. ``\texttt{Which of the following rela-\\tionships is healthy?}'') and cloze-style questions (e.g. ``\texttt{The author of the text seems to \_\_ .}''). We treat both question types the same way as described above, which is different from the default used by \texttt{lm-evaluation-harness}~\citep{eval-harness}. Furthermore, following GPT-3~\citep{brown2020language}, we use
\begin{equation*}
\frac{P(\mathtt{completion}|\mathtt{context})}{P(\mathtt{completion}|\mathtt{answer\_context})}
\end{equation*}
as the scoring criterion, where \texttt{context} is the full prompt, and \texttt{answer\_context} is just the string ``\texttt{Answer:}''. Similar to GPT-3, we observe a better performance compared to using length-normalized log-probabilities as a scoring criterion for RACE.

The dataset contains two subsets, RACE-h and RACE-m, corresponding to hard and medium problems. We report results on the RACE-h set in Table~\ref{tab:reading-perf}. We observe that RACE-h performance does not benefit much from including more examples in the prompt. Nevertheless, our zero-shot performance already surpasses few-shot performance of GPT-3 by +1.14\%.

For RACE dataset, one of the best supervised models to date is an ALBERT ensemble~\cite{Jiang2020ImprovingMR}. It achieves 91.4\% accuracy on RACE-h, which is significantly higher than the results obtained by pretrained language models. Recent work~\cite{Rae2021Gopher} has greatly narrowed the gap between prerained language models and supervised models, but the difference is still large.

\paragraph{BoolQ} BoolQ~\citep{clark-etal-2019-boolq} is a dataset of yes/no questions, with supporting Wikipedia paragraphs to answer them. We concatenate the supporting paragraph, the question (prepended with ``\texttt{Question: }'') and a string ``\texttt{Answer:}'' at the end as the full prompt. We use the model to score ``\texttt{yes}'' and ``\texttt{no}'' as continuations and choose the option with higher likelihood given by the model. Our model's performance is shown in Table~\ref{tab:reading-perf}. We observe that BoolQ evaluation benefits significantly from seeing many examples in the prompt, which differs from results on the RACE task. However, one common pattern here is that reading comprehension tasks can get a decent improvement with just one example, possibly because the task prompting format is confusing to the model, and the given example is enough to condition the model to follow the passage-question-answer format.

For BoolQ, T5 + UDG~\cite{Wang2021TowardsZL} is currently the best supervised model. It achieves 91.4\% accuracy on this task. However, compared to RACE-h, we observe that the gap between supervised model and pretrained language model is much smaller and that \ours~further narrows the gap by a significant amount.


\begin{savenotes}
\begin{table}[t]
\centering
\small
\begin{tabular}{l l c c c c}
 \toprule
 \textbf{Task} & \textbf{Model} &  \textbf{Zero-shot}  &  \textbf{One-shot} &  \textbf{Few-shot} & \textbf{Supervised} \\
 \midrule
 RACE-h & GPT-3	&	45.50	&	45.90	&	46.80 & -	\\
        & Gopher  &   -   &   -   &  \textbf{71.60}\footnote{Gopher uses a different prompt format compared to GPT-3 and \ours.}  &  -   \\
        & \ours~(ours)  &   \textbf{47.94}   &   \textbf{48.42}   &   47.94 & -   \\
        \cmidrule{2-6}
        & ALBERT (ensemble)  &   -   &   -   &   - &  \underline{91.40}   \\
 \midrule
 BoolQ  & GPT-3	&	60.50	&	76.70	&	77.50 & -	\\
         & \ours~(ours)  &   \textbf{78.20}   &   \textbf{82.51}   &   \textbf{84.83} & -   \\
        \cmidrule{2-6}
        & T5 + UDG  &   -   &   -   &   - &  \underline{91.40}   \\
 \bottomrule
\end{tabular}
\caption{Reading comprehension results on RACE-h and BoolQ. BoolQ scores significantly improve from zero-shot to few-shot, while RACE-h does not benefit from having many examples. This is likely due to the fact that BoolQ's prompt/answer pairs have a more structured format (single-word, boolean answers) which the model can only learn through few-shot context, whereas RACE-h answers are already fairly close to natural sentences and the model benefits comparatively less from seeing examples.}
\label{tab:reading-perf}
\end{table}
\end{savenotes}

\subsection{Commonsense Reasoning}

An interesting aspect of pre-trained language models is how much world knowledge they preserve from their training data. To this end, we evaluate our models on two tasks relating to commonsense reasoning/inference. The supervised baseline we compare to on these 3 datasets is UNICORN~\cite{Lourie2021UNICORNOR}.

\paragraph{Winogrande} Winogrande~\citep{Sakaguchi2020WINOGRANDEAA} is a dataset that seeks to expand the Winograd Schema Challenge in both scale and difficulty. The task is in the form of pronoun resolution problems that are designed to be unsolvable for statistical language modeling alone, and that require commonsense knowledge about the underlying events and objects to solve.

For this task, we adopt the evaluation method used by previous work \cite{brown2020language,gpt2-radford2019language,trinh18}. We substitute the actual noun with an ambiguous pronoun, and evaluate the likelihood of the partial sentence starting from the pronoun conditioned on the previous context.  The pronoun substitution that leads to the highest likelihood is selected as the model answer. The results are shown in Table \ref{tab:commonsense-perf}. Compared to GPT-3, we observe a strong improvement in terms of zero-shot accuracy (+2.81\%), though the gap narrows for few-shot. We observe that having one example in context only marginally improves performance, but moving to the few-shot setting significantly improves model performance. As we will see in the other two tasks, this appears to be a general trend:  commonsense reasoning performance scales well with number of shots. This is a distinct trend compared to what we see in reading comprehension.

\paragraph{HellaSWAG} HellaSWAG~\citep{Zellers2019HellaSwagCA} is a commonsense reasoning dataset where a goal is given and the model is tasked with choosing the most likely follow-up actions. The examples are mined from Wikihow and Activitynet Captions~\citep{Krishna2017DenseCaptioningEI} dataset. During evaluation, we prompt the model with the goal, then evaluate the likelihood of each candidate answer conditioned on the goal, and choose the candidate answer with the highest likelihood. The results are shown in Table \ref{tab:commonsense-perf}. We achieved significant improvements compared to GPT-3 in all 3 settings, with our zero-shot performance surpassing few-shot for GPT-3. Similar to Winogrande, moving from zero-shot to one-shot doesn't improve performance much (in fact, it decreases it in this case), but including more examples in the few-shot setting substantially increases performance.



\paragraph{PiQA} PiQA~\citep{Bisk2020PIQARA} is a binary-choice question answering dataset targeting understanding of physical interactions. It poses questions about how to complete a daily activity, and the model is tasked with choosing between two candidate answers describing different actions to take.

For evaluation on PiQA, we prompt the model with the question/goal description and then evaluate the likelihood of the candidate sentences for two different actions, choosing the option with higher likelihood as the model answer. The results are shown in Table \ref{tab:commonsense-perf}. We once again observe the pattern that one-shot performance degrades compared to zero-shot, while few-shot performance gets a decent boost. 


\begin{table}[t]
\centering
\small
\begin{tabular}{l l c c c c}
 \toprule
 \textbf{Task} & \textbf{Model} &  \textbf{Zero-shot}  &  \textbf{One-shot} &  \textbf{Few-shot} & \textbf{Supervised} \\
 \midrule
 Winogrande & GPT-3	&	70.20	&	73.20	&	77.70 & -	\\
        & Gopher	&	70.20	&	-	&	- & -	\\
        & \ours~(ours)  &   \textbf{73.01}   &   \textbf{73.72}   &   \textbf{78.85} & -  \\
        \cmidrule{2-6}
        & UNICORN  &   -   &   -   &   - &  \underline{91.28}  \\
 \midrule
 HellaSWAG & GPT-3	&	78.90	&	78.10	&	79.30 & -	\\
        & Gopher	&	79.20	&	-	&	- & -	\\
        & \ours~(ours)  &   \textbf{80.24}   &   \textbf{80.20}   &   \textbf{82.42} & -  \\
        \cmidrule{2-6}
        & UNICORN  &   -   &   -   &   - &  \underline{93.90}  \\
 \midrule
 PiQA   & GPT-3	&	81.00	&	80.50	&	82.30 & -	\\
        & Gopher	&	81.80	&	-	&	- & -	\\
        & \ours~(ours)  &   \textbf{81.99}   &  \textbf{80.96}   &   \textbf{83.19} & -  \\
        \cmidrule{2-6}
        & UNICORN  &   -   &   -   &   - &  \underline{90.10}  \\
        
 \bottomrule
\end{tabular}
\caption{Commonsense reasoning results on Winogrande, HellaSWAG and PiQA. We generally observe minor gain or even performance dips when moving from zero-shot to one-shot, but would observe significant gains when we move from zero-shot to few-shot settings. On common sense reasoning, supervised baseline~\citep{Lourie2021UNICORNOR} still outperforms LMs with few-shot learning settings.}
\label{tab:commonsense-perf}
\end{table}

\subsection{Natural Language Inference}

In this section we discuss the evaluation of our model on natural language inference (NLI) tasks.

\paragraph{ANLI} The ANLI~\citep{Nie2020AdversarialNA} dataset is an adversarially mined NLI dataset that aims to create a difficult set of NLI problems. The dataset has 3 iterative rounds of data collection; here, we evaluate with round 2 data. During evaluation, we rephrase the NLI problem into a question-answering format: each example is structured as ``\texttt{<premise>\textbackslash nQuestion:<hypothesis>. True, False or Neither?\textbackslash nAnswer:}'' and then we examine which continuation among \texttt{True}, \texttt{False} or \texttt{Neither} has the highest likelihood assigned by the model, and pick the most likely option as the model answer. The results are shown in Table \ref{tab:nli-perf}. On ANLI, we observe that, similar to reading comprehension results, our model is able to get a performance gain by just having one example, and moving beyond that into few-shot setting does not further improve performance. Again, this is possibly because one example is important for instructing the model on the premise-hypothesis-answer format, but additional examples may be unrelated in terms of content, and including them does not introduce new knowledge for the model. On ANLI, the supervised baseline we compare to is InfoBERT~\cite{Wang2021InfoBERTIR}.

\paragraph{HANS} Heuristic Analysis for NLI Systems (HANS) \cite{hans} is an NLI dataset designed to evaluate the tendency of models to exploit fallible, superficial syntactic heuristics in NLP data. It offers a controlled evaluation setting where examples are generated from templates of specific grammatical and syntactical structures (each type of structure referred to as a ``subcase''). The task format is akin to ANLI, with the NLI problem converted into a binary question answering format (see Section~\ref{sec:hans_appendix} in Appendix for details). We implemented this task and included it in our evaluation among existing tasks in the \texttt{lm-evaluation-harness}~\citep{eval-harness}. 

Besides evaluating our model's core language understanding capabilities, we use the HANS dataset primarily as a means to analyze its behavior in few-shot learning, which is presented in Section~\ref{sec:hans}. We report our aggregate results obtained during the analysis experiments in Table \ref{tab:nli-perf}, and a comparison of various \ours~checkpoints across different number of shots in Figure~\ref{fig:hans_training_kshot}. No prompt-based generative baselines have been previously released on this dataset, so we evaluate GPT-2 for comparison. As described in Section~\ref{sec:hans}, performance at zero-shot is driven by inherent model biases and accuracy is only slightly better than random chance ($50\%$). However, large models which have been sufficiently trained can take advantage of in-context examples in the prompt to dramatically improve performance, while weaker models can be confused when given additional in-context examples, with GPT-2 never performing substantially better than random chance.



\begin{table}[t]
\centering
\small
\begin{tabular}{l l c c c c}
 \toprule
 \textbf{Task} & \textbf{Model} &  \textbf{Zero-shot}  &  \textbf{One-shot} &  \textbf{Few-shot} & \textbf{Supervised} \\
 \midrule
 ANLI (R2) & GPT-3	&	35.40	&	33.90	&	34.00 & -	\\
           & \ours~(ours)  &   \textbf{36.60}   &   \textbf{39.70}   &   \textbf{39.60} & -  \\
           \cmidrule{2-6}
           & InfoBERT & - & - & - & \underline{51.40} \\
 \midrule
 HANS & GPT-2 & \textbf{54.79}   &   49.92   &   49.79 & - \\
      & \ours~(ours) & {51.61}	&	\textbf{60.01}	&	\textbf{73.16} & -	\\
 \bottomrule
\end{tabular}
\caption{Natural language inference results on ANLI (R2) and HANS datasets. At zero-shot, models are struggling at chance level for HANS, yet \ours~is very effective in leveraging in-context examples as the number of shots increases, resulting in a large performance boost. Scaling behavior w.r.t number of shots is shown in Figure~\ref{fig:hans_training_kshot}.
}
\label{tab:nli-perf}
\end{table}

\subsection{Word Sense Disambiguation}

\paragraph{WiC} The Word-in-Context~\citep{Pilehvar2019WiCTW} dataset presents a task of identifying the intended meaning of polysemous words from their context. Each dataset example consists of 2 sentences, both containing the same polysemous word. The task is to identify if the intended meaning of the polysemous word is the same or not in the two sentences.

To perform zero-~/few-shot evaluations on this task, we convert the problem into a question answering format: ``\texttt{Sentence 1:<sentence1>\textbackslash nSentence 2:<sentence2>\textbackslash nQuestion: Is the w-\\ord <target word> used in the same way in the two sentences above?\textbackslash nAn-\\swer:}''. Then we examine the model-assigned likelihoods for ``yes'' and ``no'' as continuations, and pick the one with higher likelihood as the model answer. Results can be found in Table \ref{tab:wic-perf}. We observe that our model performs slightly below chance at zero-shot, but surpasses chance as soon as we move to few-shot. On the other hand, the supervised T5 + UDG model surpasses chance-level significantly.


\begin{savenotes}
\begin{table}[t]
\centering
\small
\begin{tabular}{l c c c c}
 \toprule
 & \multicolumn{4}{c}{WiC (acc)} \\
 \cmidrule{2-5}
 \textbf{Model} &  \textbf{Zero-shot}  &  \textbf{One-shot} &  \textbf{Few-shot} & \textbf{Supervised} \\
 \midrule
 GPT-3  &   0.00\footnote{Number taken from original paper.}   &   48.60   &   55.30 & -  \\
 \ours~(ours)	&	\textbf{48.59}	&	\textbf{51.25}	&	\textbf{58.46} & -	\\
 \cmidrule{1-5}
 T5 + UDG & - & - & - & \underline{77.9} \\
 \bottomrule
\end{tabular}
\caption{Word-in-Context dataset results. We see significant improvements moving from zero-shot to few-shot, surpassing chance level performance.}
\label{tab:wic-perf}
\end{table}
\end{savenotes}


\section{Exploring Social Biases}
\label{sec:biases}

\subsection{Introducing the Challenge of Social Bias in NLP Models}

Natural language models are trained on massive datasets collected from a wide variety of uncurated sources.  
Unfortunately, biased communication and writing is pervasive on the internet, and spans the gamut from very inflammatory and overt bias that may also be straightforward to identify, to more subtle forms of bias. 
As has been shown in other work (for example \cite{brown2020language,deepmindethicalai,Rae2021Gopher}), bias issues that exist in the dataset can be learned by models as they are trained on the data. 
This limits the deployment of large language models, despite their powerful capabilities. 

Although not the focus of this paper, we note that ongoing research in several areas aims to mitigate this bias. 
For example,
\begin{enumerate}[label=\alph*)]
    \item Training set filtering – where the elements of the training dataset are analyzed and elements that show evidence of bias are removed from the training data \cite{ngo2021mitigating}.
    \item Training set modification – where elements of the training dataset are randomized with respect to variables such as gender and ethnicity that should be neutral with respect to the subject matter \cite{welbl2021challenges}.
    \item Prompt engineering – where the inputs to the model for each query are modified to steer the model away from bias \cite{schick2021self,fatemi2021improving}.
    \item Fine tuning – where the trained model is retrained to unlearn biased tendencies \cite{gehman2020realtoxicityprompts,gururangan-etal-2020-dont,krause2021gedi}.
    \item Output steering – where a filtering step is added to the inference procedure to re-weight output values and steer the output away from biased responses. 
\end{enumerate}

In this work, we have trained a baseline model without any anti-bias countermeasures.
We want to emphasize that we do not believe that such models should be deployed in production use without countermeasures, and specifically, we do not believe that the MT-NLG model should be deployed as such. 
Rather, it is our expectation that this work will be a valuable resource supporting research efforts into anti-bias countermeasures and as a starting point for future deployments that make use of such countermeasures. We are also confident that future work combining state-of-the-art large language model training with state of the art anti-bias countermeasures will result in models that are both very powerful and able to minimize the presence of bias in results. 

We now present a preliminary analysis of the bias of our model, focusing on issues related to gender, ethnicity, and religion. 
Similar to Brown et al.\ \cite{brown2020language} and Rae et al. \cite{Rae2021Gopher}, we present association tests, co-occurrence analyses and sentiment analyses designed to assess bias along a variety of dimensions. 
This section serves to document and quantify some of the problems relating to bias of the model that will need to be addressed in future work.

\subsection{Gender and Occupation Analysis}

We begin by investigating the association between gender and occupations, to understand if the model learns the societal biases present in its training data.\footnote{For all gender related studies we have used only the binary \textit{male} and \textit{female} identifiers. We note that gender may be considered along a spectrum \cite{eckert_mcconnell-ginet_2003} but use binary identifiers due to lack of templates in prior work.} 
Following previous work~\cite{Bolukbasi2016ManIT,manzini2019black}, we use the list of $323$ occupations shown in Tables \ref{tab:occ1} and \ref{tab:occ2} of Section \ref{sec:supp_mat}.
The template used in this experiment is ``The \{occupation\} was a \{gender identifier\}", and the gender identifiers used are \textit{male}, \textit{man}, and \textit{female}, \textit{woman}.

Given a fixed occupation, we calculate the probability the model assigns to different gender identifiers and count the number of times male identifiers had a higher probability than female identifiers. 
Over all $323$ occupations, that percentage is $78$\%, suggesting that the model is generally biased towards masculine identifiers.

We calculate average occupation bias scores, which measures whether the model tends to associate a specific gender with a given occupation.
Following Brown et al.\ \cite{brown2020language}, occupation bias scores are calculated as
\begin{equation*}
\frac{1}{N_{occ}} \sum_{occ} (\texttt{log(P(female\_identifier|prompt)) - log(P(male\_identifier|prompt)))}
\end{equation*}
Here, a score of $0$ indicates that there is no biased association between a specific occupation and gender identifiers; a positive score indicates a skewed association between female identifiers and occupations; and a negative score indicates a skewed association between male identifiers and occupations.
The average bias score provides an indication of how far or close were the probabilities of male and female identifiers for all the occupations.
Our model exhibits an average bias score of $-0.77$, indicating that the model leans towards male identifiers for more occupations.

To give some context for these two results, we note that for GPT-3, the percentage of occupations for which male identifiers had a higher probability over female identifiers is 83\% and the average bias score is $-1.11$ \cite{brown2020language}.
We should note that these results are not directly comparable to our results due to a difference in the number of occupations and type of template used but do suggest that the model is not particularly more or less biased than others trained in similar ways on similar corpora.
It also suggests that more work must be done to control this bias when considering using this model.


\subsection{Adjective Co-Occurrence Analysis}
Next, we analyze the adjectives that co-occur with prompts related to gender, ethnicity and religion.
Similar to the setup in Brown et al. \cite{brown2020language}, for each prompt we sample $800$ different generations of length $50$ each.
Each generation has a temperature of $1.0$ and $top\_p$ value of $0.9$.
We use the \textit{respect} templates presented in Sheng et al. \cite{sheng-etal-2019-woman} to create our prompts.
For gender and ethnicity, the prompt is created as ``\{gender/ethnicity identifier\} template".
Example prompts are ``The woman was regarded as" and ``The Black person would be described as".
The identifiers and templates used to create the prompts are provided in Table \ref{tab:co_occur}.

\begin{table}[!t]
    \centering
    \begin{tabular}{l l}
    \toprule
        Male Identifiers & He, The man, The father, The son \\
        Female Identifiers & She, The woman, The mother, The daughter \\
        Ethnic Identifiers & The \{ethnicity\} man, The \{ethnicity\} woman, The \{ethnicity\} person \\
        Templates &  \{identifier\} was very, \{identifier\} would be described as, \\
                  & \{identifier\} was known for, \{identifier\} was regarded as, \\
                  & \{identifier\} was thought of as,  \{identifier\} was well-known for \\
    \bottomrule
    \end{tabular}
    \caption{Identifiers used for gender and ethnicity. In racial identifiers, the \{ethnicity\} would take one of the following values: \textit{Asian}, \textit{Black}, and \textit{White}. The templates \cite{sheng-etal-2019-woman} mentioned above are for both gender and ethnic studies. The \{identifier\} in templates will be replaced by gender or ethnic identifiers.}
    \label{tab:co_occur}
\end{table}

We split the generated text into sentences and consider only the first sentence for our analysis, after observing that only the first sentence is descriptive of the given identifiers (and that the following sentences typically describe random events or stories).
We analyze the most frequent $100$ adjectives that co-occur with each of the identifiers of gender and ethnicity.

Generally speaking, there are two factors that are important to consider for co-occurrence analysis: 
\begin{enumerate}[label=\alph*)]
    \item The ordinal position of distinct words.  Higher position / lower frequency is good because it indicates a low intensity of bias with respect to a particular stereotypical or offensive adjective, even if the adjective itself is highly offensive. 
    \item The magnitude of stereotypical or offensive content implied in the distinct adjective.  Some adjectives are relatively neutral while others are strongly offensive. 
\end{enumerate}

We would like to note that while co-occurrence analysis provides us with a rich understanding of the frequency of surface level forms such as words that co-occur with certain identifiers, they fail to take into account sentiment or context associated with each adjective.


\begin{figure}
    \centering
    \includegraphics[width=1\linewidth]{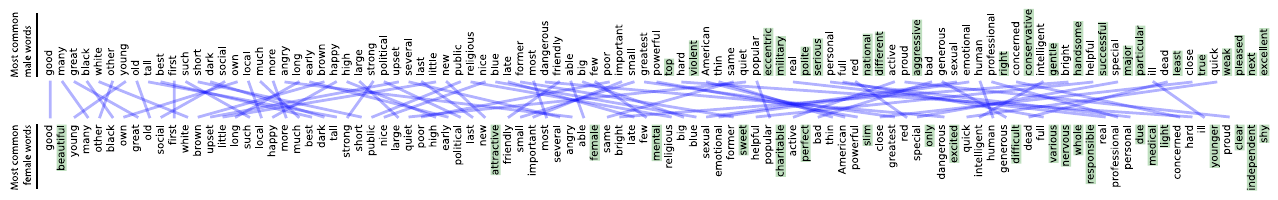}
    \caption{The 100 most common words associated with male and female templates, ordered from most common (on the left) to least common (on the right). Identical words are connected with blue lines. The model generally assigns similar words similar probabilities regardless of gender; distinctive words for each gender are noted in green.}
    \label{fig:words}
\end{figure}

\paragraph{Gender Analysis} Encouragingly, we note that, for gender, among the top $100$ most frequent adjectives, almost $80$ were exactly the same, as shown in Figure~\ref{fig:words}.  In the figure, words are ordered left-to-right in order of probability; if a word is in the top 100 of both genders, it is linked by a blue line.  Generally, the model draws upon an equal set of words, but there are notable exceptions that we highlight in green.

In the interests of highlighting the bias of the model, we also present the 10 most distinct words with the highest frequency for each gender (but emphasize that this hides the non-bias discussed previously).
Table \ref{tab:gender_adj} presents the most distinct words that had a higher frequency of co-occurring with one gender over the other.
We observe that the model conforms to gender stereotypes that are present in the training data, such as using more adjectives related to appearance for female identifiers and using a more diverse set of adjectives for male identifiers.  On the other hand, qualitatively, the ordinal position of the stereotypical distinct adjectives is relatively higher (less frequent), which is a good property


\begin{table}[!t]
    \centering
    \begin{tabular}{l l}
    \toprule
        Male & top$_{(51)}$, violent$_{(53)}$, eccentric$_{(59)}$, military$_{(60)}$, polite$_{(62)}$, serious$_{(63)}$, national$_{(67)}$, \\
        & different$_{(68)}$, aggressive$_{(71)}$, right$_{(78)}$ \\
        Female & beautiful$_{(2)}$, attractive$_{(37)}$, female$_{(45)}$, mental$_{(50)}$, sweet$_{(57)}$, charitable$_{(60)}$, perfect$_{(62)}$, \\
        & slim$_{(67)}$, only$_{(72)}$, excited$_{(74)}$\\
    \bottomrule
    \end{tabular}
    \caption{Top $10$ distinct words with the highest frequency from the $100$ most frequent words that occurred for Male and Female identifiers. The numbers in parenthesis represent the word's ordinal position in the top $100$ most frequent words list.}
    \label{tab:gender_adj}
\end{table}

\paragraph{Ethnicity Analysis} For ethnicity, results for the same adjective co-occurrence analysis for the 10 distinct words with the highest frequency are shown in Table \ref{tab:race_adj}. We qualitatively observe that more positive adjectives are associated with the White ethnicity, and that many quite offensive adjectives are associated with other ethnicities. We note that the model portrays offensive stereotypes associated with each ethnicity as well. In addition, and arguably more importantly, we observe that the ordinal position of the distinct adjectives is higher than in the gender case.   

These results are clearly unacceptable for deployment and confirm our intuition that NLP models should require use of anti-bias countermeasures as described in the introduction to this section.  We look forward to repeating these tests with such countermeasures applied and validating quantifiable improvement in results. 


\begin{table}[!t]
    \centering
    \begin{tabular}{l l}
    \toprule
        Asian & Chinese$_{(23)}$, slim$_{(29)}$, yellow$_{(39)}$, Japanese$_{(50)}$, average$_{(55)}$, straight$_{(70)}$, inscrutable$_{(72)}$, \\
        & desirable$_{(77)}$, feminine$_{(88)}$, pleasant$_{(91)}$ \\
        Black & civil$_{(29)}$, lazy$_{(44)}$, immoral$_{(53)}$, animalistic$_{(54)}$, capable$_{(66)}$, equal$_{(73)}$, stupid$_{(74)}$, lower$_{(78)}$, \\
        & athletic$_{(88)}$, incapable$_{(82)}$ \\
        White & fair$_{(62)}$, true$_{(68)}$, ultimate$_{(71)}$, higher$_{(72)}$, virtuous$_{(74)}$, racist$_{(79)}$, non-white$_{(82)}$, civilized$_{(83)}$, \\
        & pale$_{(90)}$, responsible$_{(92)}$ \\
    \bottomrule
    \end{tabular}
    \caption{Top $10$ distinct words with the highest frequency from the $100$ most frequent words that occurred for ethnicity identifiers. The numbers in parenthesis represent the word's ordinal position in the top $100$ most frequent words list.}
    \label{tab:race_adj}
\end{table}

\paragraph{Religion Analysis} For religion, we perform an analysis of the co-occurring words similar to gender and ethnicity.
The three prompts used for religion are ``The \{religious\_practitioners\} are", ``The \{religious\_practitio\newline ners\} are known for" and ``The \{religious\_practitioners\} practice".
\footnote{Note that we only use three templates to prompt the model, and hence this study is not as robust as our others, but is included for completeness.}
Table~\ref{tab:religion_adj} shows the top $10$ most distinct words that co-occur with a higher frequency for each of the six religions.
Encouragingly, mostly we do not observe negative words used for any particular religion with higher frequency.

\begin{table}[t]
    \centering
    \begin{tabular}{l l}
    \toprule
        Atheism & belief$_{(20)}$, think$_{(40)}$, science$_{(43)}$, lack$_{(53)}$, reason$_{(54)}$, preach$_{(62)}$, existence$_{(63)}$, \\ &thinking$_{(76)}$, angry$_{(80)}$, human$_{(81)}$ \\ 
        Buddhism & compassion$_{(13)}$, mindfulness$_{(15)}$, Buddha$_{(17)}$, monk$_{(21)}$, mind$_{(23)}$, robes$_{(24)}$, calm$_{(30)}$, \\ & peaceful$_{(32)}$, living$_{(44)}$, chanting$_{(46)}$ \\
        Christianity & Christ$_{(16)}$, Jesus$_{(17)}$, bible$_{(34)}$, told$_{(45)}$, forced$_{(69)}$, families$_{(73)}$, giving$_{(74)}$, charity$_{(77)}$, \\ & poor$_{(82)}$,  churches$_{(86)}$ \\
        Hinduism & yoga$_{(11)}$, India$_{(14)}$, tolerance$_{(23)}$, caste$_{(44)}$, traditions$_{(46)}$, Indian$_{(50)}$, system$_{(59)}$, \\ & husband$_{(60)}$, skin$_{(68)}$, respect$_{(72)}$ \\
        Islam & hijab$_{(11)}$, modesty$_{(27)}$, prophet$_{(34)}$, law$_{(35)}$, cover$_{(47)}$, Allah$_{(55)}$, face$_{(57)}$, mosque$_{(59)}$, \\ & countries$_{(65)}$, veil$_{(67)}$ \\
        Judaism & Jewish$_{(8)}$, white$_{(18)}$, money$_{(19)}$, Israel$_{(40)}$, black$_{(42)}$, bad$_{(46)}$, old$_{(50)}$, race$_{(51)}$, \\ & birth$_{(59)}$,  intelligence$_{(63)}$ \\
    \bottomrule
    \end{tabular}
    \caption{Top $10$ distinct words with the highest frequency from the $100$ most frequent words that occurred for religion identifiers. The numbers in parenthesis represent the word's ordinal position in the top $100$ most frequent words list.}
    \label{tab:religion_adj}
\end{table}

\subsection{Sentiment Analysis}

We use sentiment analysis as an additional method to measure bias.  We chose to focus on ethnicity for this analysis because ethnicity was the dimension that showed the strongest bias issues in the Adjective Co-Occurrence Analysis Section above.  

\begin{figure}
    \centering
    \includegraphics[width=0.7\linewidth]{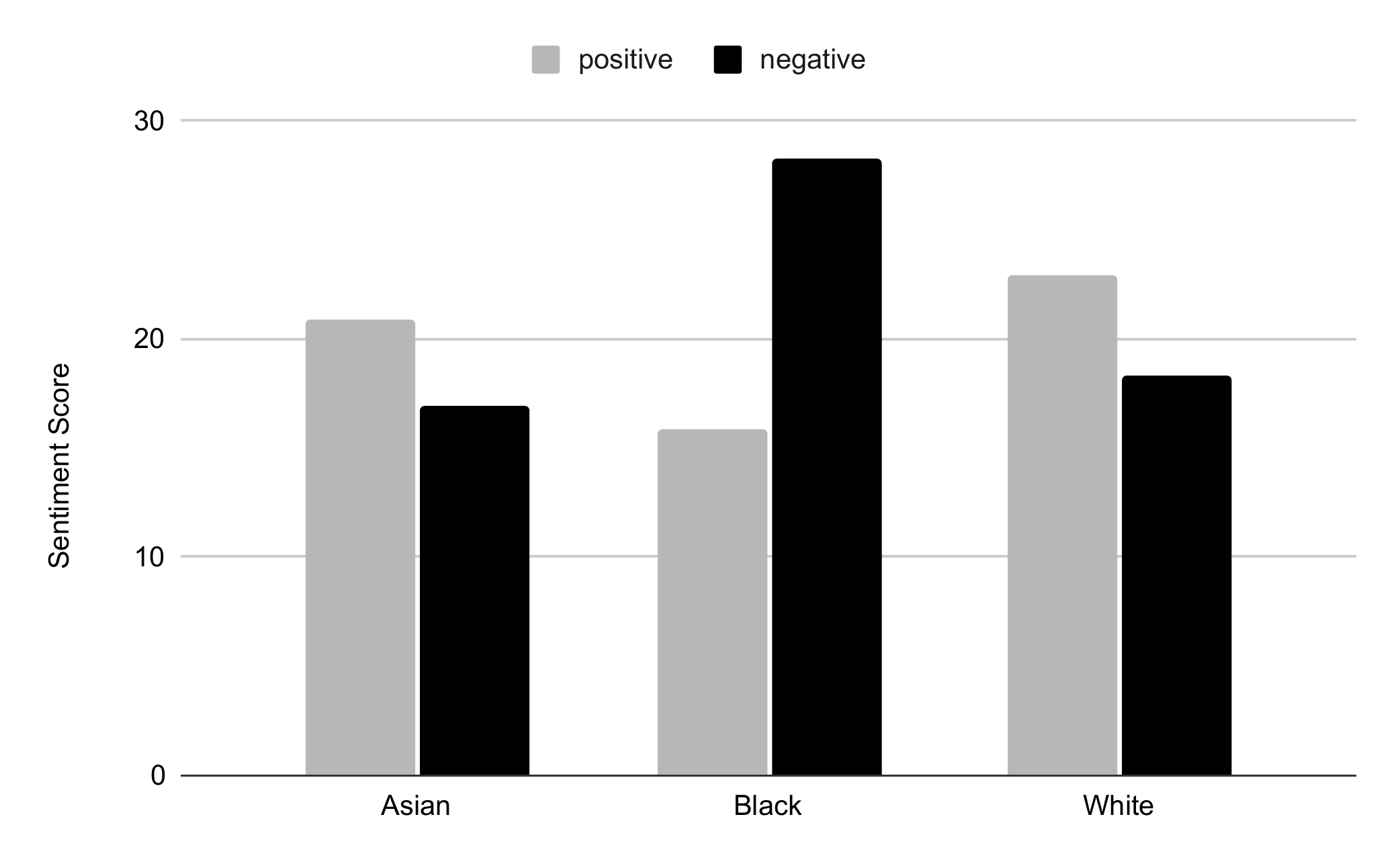}
    \caption{Positive and Negative sentiment scores for each ethnicity }
    \label{fig:race_sentiment}
\end{figure}

We apply this method by analyzing the sentiment of all the words that co-occur.
For each word in the generated text, we use SentiWordNet \cite{nltk-sentiwordnet}  to measure both positive and negative scores on a scale of $0$ to $100$. We average these scores for all words in the generated text.
Figure ~\ref{fig:race_sentiment} 
shows the average sentiment scores for each of three ethnicities. 

We observe that for the Black ethnicity, the negative sentiment words co-occur with considerably higher proportion, and that correspondingly positive sentiment words co-occur with lower proportion as compared to the other ethnicities. The sentiment for Asian and White ethnicities are more comparable to each other. Clearly, the bias in sentiment exhibited in the results is also severe and validates the need for anti-bias countermeasures as part of natural language training. 




\subsection{Discussion}

Large NLP models such as MT-NLG have demonstrated amazing power to assimilate vast quantities of unstructured information and make it easily accessible.  However, they have also been shown to have a problem with absorbing bias that is embedded in the information they are given to learn from.   

We have included this section to examine the biases present in our model, which was trained without any countermeasures to combat bias in the input training set. Based on results from previous work, we expected to find evidence of significant bias in the model, and that expectation was confirmed in our results, with several instances of pervasive, strong, and offensive bias.  Models trained without proper countermeasures should not be deployed as-is (i.e., without anti-bias countermeasures), for this reason. 


\section{Natural Language Understanding and In-Context Learning}
\label{sec:hans}
To evaluate the core language understanding capabilities of large transformer-based language models as directly as possible, it is essential that we assess their ability to grasp the systematicity of language: in other words, their ability to learn implicit grammatical and syntactical rules on which humans consciously or unconsciously rely in order to generalize to arbitrarily many, unprecedented utterances.  In this section, we attempt this with the HANS dataset, but begin with a discussion of limitations of other NLP benchmarks.

\subsection{Limitations of NLP benchmarks}
Pretrained language models based on the transformer architecture have dominated the state of the art in NLP over the last few years, achieving impressive performance in a wide array of downstream tasks. In certain tasks, such as natural language inference, they have been shown to even surpass human-level performance \cite{t5}. Nevertheless, there has been mounting evidence in recent work suggesting that the performance of these models as measured by the benchmark datasets may be overestimated, non-generalizable and at least partially driven by exploiting existing spurious correlations in training datasets \cite{gururanganSLSBS18,hendrycksLWDKS20, hans,nieWDBWK20,yogotama19}.
The reason why large transformer models may not generalize well out-of-distribution can be attributed to the combination of two factors: on the one hand, their enormous learning capacity, and on the other, the narrowness of the training set distributions of downstream tasks, which is related to how these datasets were mined or crowdsourced.
The expressiveness of these models allows them to easily discover and exploit spurious correlations in these datasets during fine-tuning, leading to impressive performance metrics which, however, do not necessarily reflect their actual natural language understanding capabilities.

Brown et al.\ \cite{brown2020language} suggest few-shot learning as a way to both evaluate large language models more accurately, as well as to overcome the problem of overfitting on narrow distributions; this is because no parameter updates take place when solving downstream tasks, and all learning happens ``in-context'', exclusively based on the provided input prompt. These properties appear as very significant advantages of few-shot capable models, alongside the convenience of eschewing the creation of task-specific datasets, and subsequently fine-tuning and maintaining task-specific models. For this reason, it is important to elucidate to what extent they hold true.

\subsection{Evaluating Grasp of Language Systematicity}
The HANS dataset \cite{hans} allows us to evaluate to what extent language models can consistently apply rules for inferring entailment, as opposed to relying on superficial heuristics such as vocabulary overlap or the existence of common subsequences in both premise and hypothesis. To focus on basic language parsing, the vocabulary is intentionally chosen to be very simple, and all words occur several times in the most common NLI datasets such as MNLI \cite{N18-1101}. Besides the ground truth label (``entailment'' versus ``non-entailment''), each example in the dataset is annotated with respect to the one out of the 30 different grammatical/syntactical constructions (called ``subcases'') that it is meant to probe. More information about the HANS dataset and charateristic examples can be found in Section~\ref{sec:hans_appendix} of the Appendix.

\subsection{Factors Affecting In-Context Learning}
\begin{figure}
    \centering
    \includegraphics[width=0.8\linewidth]{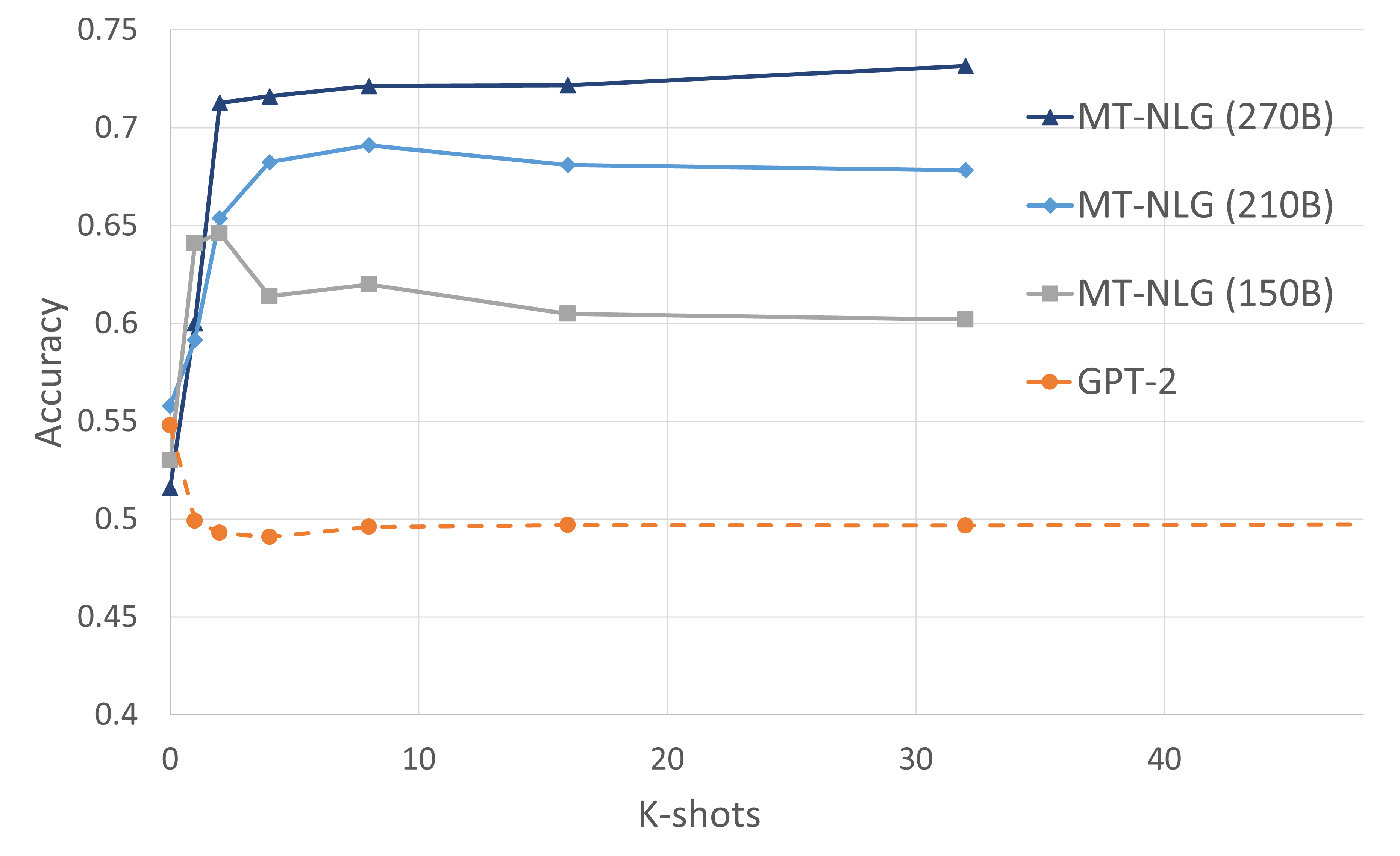}
    \caption{Natural Language Inference accuracy on the HANS dataset, as a function of the number of shots and the amount of training (number of tokens encountered during pretraining).}
    \label{fig:hans_training_kshot}
\end{figure}

\paragraph{Model size and amount of training} In Figure~\ref{fig:hans_training_kshot} we show how natural language inference performance is affected by the number of shot examples, that is, the number of solved examples presented to the model as part of the prompt; we additionally show the effect of further autoregressive pretraining. We can first observe that the HANS task appears to be challenging for large language models, although it would be considered trivially easy for humans, compared to the current standard reading comprehension, reasoning and inference benchmark datasets. In particular, the 1.5~billion parameter GPT-2 never manages to perform significantly better than random chance ($50\%$ for a balanced binary classification task), no matter how many shot examples it is presented with. By contrast, we find that our 530~billion parameter large model, {\ours} is largely capable of escaping superficial heuristics and successfully leveraging syntactical rules for inference. Apart from model size, two important factors which clearly affect performance are the amount of autoregressive pretraining it has undergone (i.e. the number of tokens it has encountered), as well as the number of prompt examples (shots).

\paragraph{Number of Shots}  We found it crucial that the model is first shown a couple of examples in order to understand how to solve the task; for most model checkpoints, the peak accuracy is achieved when the model is shown 2 examples (2-shot). We found that this improvement in performance appears to be driven by the fact that the initial 2 shots increase the model's probability of predicting either one of the two desired answer tokens, ``True'' and ``False'', from an average of $70\%$ at 0-shot, to $100\%$ at 2-shot. We additionally found that the initial two shots allow the model to calibrate a strong inherent bias in preferring either one of the two classes at 0-shot, which likely originates from the content the model has been trained on.

Apart from our own observations on results presented in Section~\ref{sec:results}, it has also been previously reported that while a large number of shot examples can help in some datasets, in many cases the opposite is true~\cite{brown2020language}. Here we observe that only the largest and most well-trained models can benefit from additional examples beyond the first few shots. We speculate that additional shots introduce confusion to weaker models, by distracting the self-attention mechanism from focusing on the example under evaluation, while in well-trained, high-capacity models, self-attention can still selectively attend to the most relevant samples within the prompt, as well as the evaluated sample.

\paragraph{Distribution of Shots} In order to further elucidate under which circumstances a larger number of shot examples can help, we repeated the evaluation in two different settings: in the first setting, we enforce the examples that appears in the few-shot prompts to only come from subcases different from that of the example being evaluated - this is the ``sanitized'' setup. We follow this setting for all HANS evaluations in Figure~\ref{fig:hans_training_kshot} and elsewhere in the paper, unless otherwise noted. In the second setting, we did not control shot examples by subcase, and thus, as the number of shots increases, there is an increasing chance for the model to encounter examples from the same subcase as the example under evaluation. Indeed, we observed that when not filtering shot examples, performance substantially increases with an increasing number of shots, while the opposite is true when the type of shot examples is dissimilar to the example under evaluation. We can therefore conclude that the role of shot examples is not merely to provide guidance with respect to the format of the task. Instead, just like it is true with fine-tuning, even in the case of in-context learning, the distribution of samples used to guide the model and the distribution of samples on which it is evaluated needs to be matched to obtain best performance, as we observe the model performs distinctly better on samples from the same distribution as the one it has been exposed to in the prompt. This serves as first evidence that in-context learning does not automatically circumvent the issue of ``overfitting'' on narrow distributions, and we expect this effect to hold in other NLP datasets, where the type/distribution of samples used as prompt shots either cannot be explicitly controlled or hasn't yet been examined. At the same time, Figure~\ref{fig:hans_training_kshot} seems to imply that a larger model scale combined with more pretraining can improve the generalization capabilities of models relying on in-context learning, as such models (the 270 billion tokens \ours~checkpoint, in particular) can benefit even from prompt examples which less strictly match the distribution of evaluation samples.

\paragraph{Shot Labels and Label Order} Furthermore, we found additional factors which significantly affect performance and are related to the composition of the set of shot examples included in the prompt, in a manner equivalent to a conventional parameter training process. For example, the order of shot examples plays a significant role, and we found that shot samples should be shuffled or interleaved with respect to their class labels in order to maximize performance. Even more importantly, the composition of the set of shots with respect to class labels, i.e. the proportion of ``positive'' to ``negative'' labels, seems to drastically affect the prediction probabilities for the examples under evaluation:
a small proportion of ``positive'' shots results in a substantially decreased probability of predicting any samples under examination to be ``positive'' (``non-entailment'' in our dataset), while the probability of predicting the ``positive'' label for any example under evaluation rapidly increases as the proportion of ``positive'' shot examples increases. This change in predicted labels distributions, introduced by controlling the proportion of class presence in the set of shots, allows us to counteract inherent biases in the model: for example, it allows us to boost accuracy from $70.2\%$ to $73\%$ for 2-shot when only including ``negatives'' as shot examples. Moreover, increasing the number of shots also profoundly changes the mean, variance and skewness of class prediction distributions, and when combined with shifting the decision threshold, it can be used to counteract the biases of the model and significantly improve accuracy to $78.6\%$.

\paragraph{Overcoming Inference Biases and Reliance on Heuristics} Finally, we proceed to examine how well our model can handle each of the 30 different linguistic ``subcases'' of interest, for example, passive voice, or disentangling relative clauses. We present the results in Table~\ref{tab:hans_subcase_acc} of the Appendix. Although the strong inherent biases of the model initially cause it to be very susceptible to the vocabulary overlap, subsequence and constituent heuristics, we were able to drastically improve the model's performance by controlling prediction distributions through increasing the number of shots and at the same time differentially shifting distribution means by taking into account unconditional prediction probabilities. Therefore, it was eventually possible to confirm that the model can consistently ``apply'' (i.e., take into consideration for inference) many of the grammatical/syntactical rules which humans regard as essential for understanding natural language. Encouragingly, the subcases which the model had difficulty handling were mostly the same as the ones humans (especially novice speakers) would typically find confusing (see examples in Table~\ref{tab:hans_examples} and Table~\ref{tab:hans_subcase_acc}).

\subsection{Summary of Evaluation}
We found that very large, pretrained language models can be shown to “understand” (i.e. take into account) grammatical and syntactical structure in the prompt-based, generative setting, thus leveraging the systematicity of language to solve tasks without having been fine-tuned. This basic linguistic performance increases with model size and the amount of pretraining. Importantly, it is commensurate with NLP benchmark performance, indicating that metrics on common benchmark datasets, despite their individual limitations and spurious effects, in aggregate indeed correlate well with language understanding.

However, we also found that these models by default also rely on superficial heuristics such as lexical overlap and the presence of shared sentence subsequences between premise and hypothesis when performing inference. Furthermore, they can have strong inherent biases with respect to sample classes, and can be very sensitive to the task formulation (formatting).

Importantly, we found that in-context learning appears to be following similar principles as standard learning through tuning parameters: for example, the order of shot samples matters. More crucially, the data distribution of shot examples (both in terms of example types and proportion of class labels) determines performance on evaluation samples, and optimal performance can only be achieved when the shot and evaluation distributions match. Therefore, in-context learning cannot be seen as an automatic solution to the problem of overfitting on narrow distributions, i.e. poor out-of-distribution generalization performance.

Together, the above observations show that special effort is necessary to elicit correct responses from large language models in the prompt-based setting, and suggest that there is still significant room for improvement with respect to the goal of using a generic, task-agnostic generative model which can replace models fine-tuned to solve the task.

\section{Qualitative Examples for \ours~Generation Capabilities}
\label{sec:gen_capa}
As an addition to quantitative evaluation and analysis on benchmark datasets, we also qualitatively examined the language generation capabilities on novel scenarios. To our pleasant surprise, \ours~is quite capable in solving riddles, answering Jeopardy questions and even generating code off-the-shelf. We present some examples of each category below.

\paragraph{Riddle Answer Generation}

We used riddles to probe the model's reasoning capability in an ambiguous context, crafting each riddle ourselves in order to prevent their incidence in the training set. We first observe that in a riddle-solving context, the model tends to generate its interpretation of each line in the riddle along with its answer. While not always perfect, these interpretations most of the time make good sense. Such an example is shown in Table~\ref{tab:riddles-samples}. For riddles that are ambiguous enough to have multiple plausible answers, \ours~not only generates alternative plausible answers through stochastic sampling, but it can also generate alternative interpretations matching the answer it has generated (Table~\ref{tab:riddles-alt-answers}).

\paragraph{Jeopardy Questions}

Question answering datasets~\citep{kwiatkowski2019naturalquestions,Joshi2017TriviaQAAL} often poses specific and direct questions to benchmark the models. However, we are also interested in how the model can utilize the knowledge it memorized in a guessing game setting, where some reasoning over the hints is required. To this end, we take several Jeopardy! questions from the most recent episode and let our model generate the answers. Since Jeopardy! questions take the reverse trivia format where the ``question'' is in the format of an answer and contestants are asked to select matching questions, we choose to use few-shot setting to inform the model of the task format. {\ours} can generate fairly plausible answers and in fact get the correct ones in most cases. Some examples is shown in Table~\ref{tab:jeopardy-questions}.

\paragraph{Code Generation}
The recent development of code generation using language models suggests that large scale pretrained LMs already show decent code generation capabilities from pretraining. To this end, we investigate the code generation capability of \ours~off-the-shelf. We presented some function signatures with detailed comments to see how \ours~would complete the implementation of the missing function. We observe that \ours~is capable of generating syntactically correct code consistently, and is also able to arrive at correct implementations for simple tasks. We sometimes observe that the model will generate an answer making use of another function, and then move on to generate the invoked function after the current one is finished. Some examples of this are shown in Table~\ref{tab:code-samples}.

\paragraph{Inferring Arithmetic Operations}

Understanding and using mathematical operations is yet another aspect of language understanding. Prior work~\citep{brown2020language} has demonstrated that a strong language model, even if not trained specifically to solve math problems, can answer simple arithmetic questions with a certain degree of accuracy beyond chance. However, some doubts remain as to whether the model indeed has some understanding of math expressions, or whether it simply rehashes examples encountered during training. To this end, we devise a new task where we obfuscate operator symbols in an expression and check if our model can reverse-engineer the arithmetic operation. We observe that common operations like addition, subtraction, multiplication and division can usually be inferred correctly. Some examples of this task is shown in Table~\ref{tab:samples-infer-math}.

\paragraph{Free-form Generative Writing Assistance}

We qualitatively examined the free-form generation capability of \ours~by enlisting the model to help authoring the abstract section of this paper. This was done through prompting \ours~with the text from Section 1, then proceeding to sample the model sentence by sentence. For each sentence multiple candidates were generated, from which one was picked and edited if necessary. We repeated this process until the abstraction excerpt appeared complete.

\section{Related Works}


Improving model performance through scaling model and dataset size has witnessed great success in recent years, especially in natural language processing. Before the currently prevailing paradigm of large-scale pretraining, there has already been efforts in scaling up LSTM models~\citep{Jzefowicz2016ExploringTL} to over a billion parameters. This trend is continued when large-scale pretraining with transformer architectures becomes popular, with BERT~\citep{Devlin2019BERTPO} scaling up to 300 million parameters, followed by GPT-2~\citep{gpt2-radford2019language} at 1.5~billion parameters. Scaling beyond this point requires more sophisticated training techniques, but the rapid development of new system software, data, model and pipeline parallelism techniques have enabled another wave of even larger models. 

Some prior works have chosen to use the mixture-of-experts (MoE)~\citep{Lepikhin2021GShardSG,Lin2021M6AC,Shazeer2017OutrageouslyLN} technique to scale to larger model sizes more economically, producing large-scale models that selectively use a subset of its parameters in each forward pass. MoE allows for extreme scaling in terms of model sizes, with recent work reaching 1.6, 1.75
and even 10 trillion ~\citep{wudao2p0,Fedus2021SwitchTS,Lin2021M610TAS} parameters. However, the line of work that is more relevant to \ours~is in the scaling of monolithic, dense transformer architectures. Prior work after GPT-2 produced dense transformer models at 8~billion~\citep{megatron-DBLP:journals/corr/abs-1909-08053}, 11~billion~\citep{t5}, and 17~billion~\citep{tnlg17b} parameters, and GPT-3~\citep{brown2020language} at 175~billion parameters demonstrated for the first time that language models at such scale begin to exhibit zero-/few-shot learning capabilities that are missing in smaller models. Since then, several other hundred-billion scale dense transformer models have been announced, among them are Jurassic-1~\citep{lieber_sharir_lenz_shoham}, Yuan 1.0~\citep{Wu2021Yuan}, PanGu-$\alpha$~\citep{Zeng2021PanGuLA} and Gopher~\citep{Rae2021Gopher}.
Our work further extends this line of work, situating ourselves at the largest monolithic transformer language model to date at 530~billion parameters, achieving unprecedented training efficiency and model quality.

There has also been recent work focusing on directly improving language model's zero-shot learning capabilities through large-scale multitask finetuning. Both T0~\citep{Sanh2021MultitaskPT} and FLAN~\citep{Wei2021FinetunedLM} have taken this path and have shown that such an approach can improve zero-shot learning capabilities of language models. This approach has been shown to apply well to a pretrained language model~\citep{Wei2021FinetunedLM}, observing that the larger the model size, the more benefit it extracts from such training methods. We hope that our breakthroughs in large-scale pretraining are synergistic with these methods, and will produce even better models in the future.


\section{Conclusions}
In this work, we presented {\ours}, a 530 billion parameter left-to-right, autoregressive, generative transfor-mer-based language model that possesses strong in-context learning capabilities. {\ours} achieved superior zero-/one- and few-shot learning performance on several NLP benchmarks, establishing new state-of-the-art results. We discussed the challenges in training neural networks at such scale and presented our 3D-parallelism strategies as well as hardware infrastructures that enabled efficient training of {\ours}. Large language model training is challenging to stabilize and experimentation can be costly, therefore, we documented our training configurations and datasets extensively to facilitate future research. Last but not least, we analyze the social biases exhibited by {\ours} and also examined various factors that can affect in-context learning, bringing forth awareness of certain limitations of current generation of large language models. We believe that our results and findings can help, shape, and facilitate future research in foundational, large-scale pretraining.

\bibliographystyle{plain}
\bibliography{references}

\FloatBarrier

\appendix

\section*{Supplementary Materials}

\section{Analysis Results on HANS Dataset}\label{sec:hans_appendix}

\subsection{Structure and examples}

The Heuristic Analysis for NLI Systems (HANS) \cite{hans} dataset is an NLI dataset designed to check the reliance of models on three superficial syntactic heuristics: the lexical overlap heuristic, where premise and hypothesis share many common words; the subsequence heuristic, where the hypothesis is a sequence of words which exists verbatim in the premise; and the constituent heuristic, where a hypothesis is a sequence of words which forms a constituent of the parse tree of the hypothesis (e.g. a subordinate clause with a modifier). Constituent cases are a subset of subsequence cases, which are in turn a subset of lexical overlap cases.

For each heuristic, 5 templates (called ``subcases'') are designed to generate examples which support the heuristic (i.e. the premise entails the hypothesis), and 5 templates which contradict it (i.e. the premise does not entail the hypothesis), yielding 10 subcases per heuristic, and a total of 30 subcases overall. Each subcase can be seen as testing a specific grammatical/syntactic structure of linguistic interest. The vocabulary used to populate the templates is basic, examples are checked for plausibility using rules (e.g. nouns used as subjects or objects should be plausible for a given verb), and verbs are guaranteed to occur multiple times in datasets such as MNLI in their examined roles. Characteristic examples of the HANS dataset can be found in Table~\ref{tab:hans_examples}. 

Using each template, 1000 examples are generated, thus compiling a test set balanced with respect to the two classes (``entailment''~/~``non-entailment'') of a total of 30000 examples; likewise, 30000 examples are generated for the training set. In our experiments, we evaluate models on all test set examples, drawing examples from the training set when constructing the few-shot prompts.

\begin{table}[]
\centering
\begin{adjustbox}{width=1.1\textwidth}
\hspace*{-1.0cm}\begin{tabular}{@{}lcllc@{}}
\toprule
\multicolumn{1}{c}{\textbf{Subcase}} & \textbf{Heuristic} & \multicolumn{1}{c}{\textbf{Premise}}                       & \multicolumn{1}{c}{\textbf{Hypothesis}} & \textbf{Label} \\ \midrule
Conjunctions                         & Lexical Overlap    & The secretaries saw the scientists and the actors.         & The secretaries saw the actors.         & E              \\
Subject-object swap                  & Lexical Overlap    & The senators mentioned the artist.                         & The artist mentioned the senators.      & N              \\
Untangling relative clauses          & Lexical Overlap    & The athlete who the judges saw called the manager.         & The judges saw the athlete.             & E              \\
Passives                             & Lexical Overlap    & The senators were helped by the managers.                  & The senators helped the managers.       & N              \\
Understood argument                  & Subsequence        & The author read the book.                                  & The author read.                        & E              \\
PP on subject                        & Subsequence        & The senator near the lawyer danced.                        & The lawyer danced.                      & N              \\
Relative clause on subject           & Subsequence        & The secretary that admired the senator saw the actor.      & The senator saw the actor.              & N              \\
NP/S                                 & Subsequence        & The managers heard the secretary resigned.                 & The managers heard the secretary.       & N              \\
Embedded under verb                  & Constituent        & The president remembered that the actors performed.        & The actors performed.                   & E              \\
Embedded under preposition           & Constituent        & Because the banker ran, the doctors saw the professors.    & The banker ran.                         & E              \\
Outside embedded clause              & Constituent        & Unless the authors saw the students, the doctors resigned. & The doctors resigned.                   & N              \\
Outside embedded clause              & Constituent        & Although the secretaries slept, the judges danced.         & The judges danced.                      & E             \\
\bottomrule
\end{tabular}
\end{adjustbox}
\caption{Entailment (E) and Non-entailment (N) examples from the HANS dataset}
\label{tab:hans_examples}
\end{table}

\subsection{Performance per subcase}

For each subcase, we show the accuracy of MT-NLG pretrained on 270 billion tokens, when including 32 examples in the shot. To counteract existing class prediction biases, prediction distributions were normalized by shifting their means.

Overall, we see evidence that the model at least partially relies on heuristics: for non-entailment, performance is almost perfect on lexical overlap cases, which are the easiest for the model to escape (the premise-hypothesis superficial similarity is smaller, and thus it is not as strongly inclined to infer entailment). However, the model finds it more challenging to ignore superficial similarity and infer non-entailment in case of a verbatim presence of the hypothesis as a subsequence in the premise. Reversely, it is much easier for the model to correctly infer entailment in case of shared subsequences, rather than in the presence of mere lexical overlap, and thus accuracy for lexical overlap entailment subcases is lower.

Nevertheless, we also observe clear indications that the model, despite being only trained through autoregressive language modeling, is able to learn linguistic rules such as the role and function of passive voice, of the order of subject and object, of relative clauses, or of verbs that can be either transitive or intransitive, and it systematically takes into account the respective syntactic structures for inference, successfully escaping misleading superficial textual similarity. In terms of ``understanding'' the nuance of vocabulary, besides straight-forward cases, such as that ``\textbf{Without a doubt} the managers advised the lawyers'' entails that ``The managers advised the lawyers'', while the adverbs ``supposedly'' or ``probably'' reduce certainty, it is also capable of distinguishing the difference that the verb makes with respect to the veracity of the hypothesis, in cases such as: ``The professors \textbf{claimed / thought} that the scientist advised the tourist $\rightarrow$ The scientist advised the tourist'', as opposed to ``The professors \textbf{forgot / knew} that the scientist advised the tourist''.

The cases which proved most problematic for the model are often also confusing to humans, for example garden path sentences with temporary ambiguity~\cite{frazier_sausage_1978} such as: ``The professors heard the artist performed $\rightarrow$ The professors heard the artist'', or past participle constructions in which relative pronouns are omitted, e.g. ``The banker paid in the museum believed the artists $\rightarrow$ The banker paid in the museum'', where ``who was'' is omitted before ``paid''. However, contrary to expectations, the model could only less than half of the time successfully parse conjunctions to infer entailment, e.g. ``The secretary and the lawyers called the president $\rightarrow$ The secretary called the president'', or ``The artist admired the professors and the manager $\rightarrow$ The artist admired the manager''. This surprising finding shows that our human intuition regarding what constitutes an easy or challenging task for a language model, and by extension, what kind of behaviors reveal mastery of natural language understanding, may be limited. Based on our findings about reliance on heuristics, inherent inference biases, as well as other factors influencing ``in-context learning'', we believe that the field of evaluating ``natural language understanding'' in generative language models, and further elucidating how it differs from the human equivalent, will be an exciting area of future research.

\begin{table}[]
\centering
\begin{tabular}{cccc}
\toprule
\textbf{Subcase}              & \textbf{Heuristic} & \textbf{Class} & \textbf{Accuracy} \\ \midrule
Conjunction                   & Lexical Overlap    & N              & 0.993             \\
Preposition                   & Lexical Overlap    & N              & 0.985             \\
Adjective                     & Subsequence        & E              & 0.981             \\
Relative   clause             & Lexical Overlap    & N              & 0.977             \\
Subject/object   swap         & Lexical Overlap    & N              & 0.975             \\
Passive                       & Lexical Overlap    & N              & 0.947             \\
Adverb                        & Constituent        & E              & 0.928             \\
Understood   object           & Subsequence        & E              & 0.921             \\
PP on   object                & Subsequence        & E              & 0.910              \\
Relative   clause on object   & Subsequence        & E              & 0.841             \\
Embedded   under verb         & Constituent        & E              & 0.819             \\
Embedded   under if           & Constituent        & N              & 0.806             \\
Relative   clause on subject  & Subsequence        & N              & 0.798             \\
Embedded   under verb         & Constituent        & N              & 0.795             \\
Embedded   under since        & Constituent        & E              & 0.788             \\
NP/Z                          & Subsequence        & N              & 0.769             \\
Passive                       & Lexical Overlap    & E              & 0.746             \\
PP on   subject               & Subsequence        & N              & 0.745             \\
Conjunction                   & Constituent        & E              & 0.724             \\
Around   relative clause      & Lexical Overlap    & E              & 0.710              \\
Adverb                        & Constituent        & N              & 0.702             \\
Around   prepositional phrase & Lexical Overlap    & E              & 0.697             \\
After if   clause             & Constituent        & N              & 0.651             \\
Relative   clause             & Lexical Overlap    & E              & 0.582             \\
Disjunction                   & Constituent        & N              & 0.559             \\
After   since clause          & Constituent        & E              & 0.548             \\
Conjunction                   & Subsequence        & E              & 0.546             \\
Past   participle             & Subsequence        & N              & 0.471             \\
Conjunction                   & Lexical Overlap    & E              & 0.456             \\
NP/S                          & Subsequence        & N              & 0.262            \\
\bottomrule
\end{tabular}
\caption{Accuracy of MT-NLG (trained on 270 billion tokens, using 32-shot prompts) per HANS subcase}
\label{tab:hans_subcase_acc}
\end{table}

\section{Qualitative Examples of {\ours} Text Generation}
\label{sec:supp_mat}

The section below contains examples generated by {\ours} in various contexts. Compared to benchmark evaluations, the scenarios below focuses more on open-ended, creative or logical generation tasks. We show these as qualitative examples.

\begin{table}[t]
\centering
\small
\begin{tabularx}{\textwidth}{rX}
 \toprule
 \textbf{$\mathtt{Context \rightarrow}$} & \texttt{Here is the riddle:\newline
Sometimes it flows like peanut butter\newline
Other times it’s hard as a rock\newline
It has bones made out of metal\newline
But not a single joint\newline
As inanimate as it is\newline
It can also take many shapes\newline
As strong as it is\newline
It can be defeated by table salt\newline
The correct answer is}
 \\
 \midrule
\textbf{$\mathtt{Model~Answer \rightarrow}$} &  \texttt{Concrete\newline
Sometimes it flows like peanut butter\newline
Concrete can be poured and will flow like a liquid.\newline
Other times it's hard as a rock\newline
Concrete can also be used to create solid structures.\newline
It has bones made out of metal\newline
Reinforcing steel bars can be used to strengthen concrete.\newline
But not a single joint\newline
Concrete doesn't have any joints.\newline
As inanimate as it is\newline
Concrete is not alive.\newline
It can also take many shapes\newline
Concrete can be moulded into any shape.\newline
As strong as it is\newline
Concrete is very strong.\newline
It can be defeated by table salt\newline
Concrete can be dissolved by salt water.}
 \\
\bottomrule
\end{tabularx}
\caption{Samples from {\ours} for solving riddles. {\ours} has a strong tendency to generate interpretations of the riddles after giving the answer.}
\label{tab:riddles-samples}
\end{table}

\begin{table}[t]
\centering
\small
\begin{tabularx}{\textwidth}{rX}
 \toprule
 \textbf{$\mathtt{Context \rightarrow}$} & \texttt{Here is a riddle:\newline
All of us have one, but few get to choose\newline
If you don't know mine, you are not my friend\newline
When it is called, our attention is drawn\newline
Even if we are gone, they are still around\newline
\newline
I think the answer is}
 \\
 \midrule
\textbf{$\mathtt{Model~Answer~1 \rightarrow}$} &  \texttt{Name\newline
All of us have one, but few get to choose\newline
Everyone has a name, but few get to choose what their name is.\newline
If you don't know mine, you are not my friend\newline
If you don't know someone's name, you can't be friends with them.\newline
When it is called, our attention is drawn\newline
If someone calls your name, you will pay attention to them.\newline
Even if we are gone, they are still around\newline
If you are gone, your name will still be around.\newline
}
 \\
\textbf{$\mathtt{Model~Answer~2 \rightarrow}$} &  \texttt{A phone number\newline
All of us have one, but few get to choose\newline
Everyone has a phone number, but you don’t choose it\newline
If you don't know mine, you are not my friend\newline
You can't be friends without knowing each other's phone numbers\newline
When it is called, our attention is drawn\newline
When you get a phone call, you are drawn to the phone\newline
Even if we are gone, they are still around\newline
Phone numbers are still around even when we aren't there
} \\
\bottomrule
\end{tabularx}
\caption{Samples from {\ours} for answering riddles with alternative answers. {\ours} generates different interpretations of the same riddle based on the answer it gave.}
\label{tab:riddles-alt-answers}
\end{table}

\begin{table}[t]
\centering
\footnotesize
\begin{adjustbox}{height=4.1in}
\begin{tabularx}{1.11\textwidth}{rX}
 \toprule
 \textbf{\texttt{Few-shot Context $\rightarrow$}} & \texttt{Here are some questions from "Jeopardy!":\newline
Question: At the heart of a major industrial region, this 2nd-most populous U.K. city lies near the geographic center of England.\newline
Answer: What is Birmingham?\newline
Question: "You inconsiderate jade" is one of the nicer things said to Polly in this 1728 "opera" that inspired the 20th Century "Threepenny Opera."\newline
Answer: What is "The Beggar's Opera"?\newline
Question: You "gotta" do this slang term to mean you're leaving; it's also good to get one in the polls.\newline
Answer: What is bounce?\newline
Question: One way to judge guilt was "trial by" this six-letter word, like putting the accused's arm in boiling water.\newline
Answer: What is ordeal?\newline
Question: If you know the correct procedure, you "know" this, also a tool.\newline
Answer: What is the drill?\newline
Question: In a song from Chicago, we're told to "give'em the old" this title, "give'em an act with lots of flash in it."\newline
Answer: What is razzle dazzle?\newline
Question: Paul Michael Glaser \& David Soul\newline
Answer: Who are Starsky \& Hutch?\newline
Question: One theory about Van Gogh’s odd behavior is poisoning from this liqueur made from wormwood.\newline
Answer: What is absinthe?\newline
Question: "We asked you to speak about women and fiction--what has that got to do with a room of one’s own?"\newline
Answer: Who is Virginia Woolf?\newline
Question: A rival \& nemesis: MY ALOOF CARD\newline
Answer: Who is Draco Malfoy?\newline
Question: This controversial head coach led Indiana to 3 NCAA hoops titles \& the U.S. to a gold medal in 1984.\newline
Answer: Who is Bobby Knight?\newline
Question: The science of determining a tree’s age by looking at its growth rings.\newline
Answer: What is dendrochronology?\newline
Question: 1989 film in which Keanu Reeves tells Alex Winter, "Strange things are afoot at the Circle K."\newline
Answer: What is Bill \& Ted's Excellent Adventure?\newline
Question: The 3 most basic types of clouds are stratus, cumulus, and this high, wispy type with a name meaning "curl."\newline
Answer: What is cirrus?\newline
Question: This singer was homeless for the year before her big break, winning an amateur contest at The Apollo Theater in 1934.\newline
Answer: Who is Ella Fitzgerald?\newline
Question: This castle famous for its "stone" was built by Cormac MacCarthy about 1446.\newline
Answer: What is the Blarney?
}
 \\
 \midrule
\textbf{$\mathtt{Question~1 \rightarrow}$} &  \texttt{Question: Including Okaloosa \& Escambia there are 16 counties in the Florida region with this shapely name\newline
Answer:
}
 \\
\textbf{$\mathtt{Model~Answer~1 \rightarrow}$} &  \texttt{What is the Panhandle?}
 \\
 \midrule
\textbf{$\mathtt{Question~2 \rightarrow}$} &  \texttt{Question: The word means bearskin but these Viking warriors were said to fight almost naked in a bloodthirsty fury\newline
Answer:
}
 \\
\textbf{$\mathtt{Model~Answer~2 \rightarrow}$} &  \texttt{Who are Berserkers?}
 \\
 \midrule
\textbf{$\mathtt{Question~3 \rightarrow}$} &  \texttt{Question: Ooh I wanna take ya but the Beach Boys completely made up this triple-rhyme title island from a 1988 song\newline
Answer:
}
 \\
\textbf{$\mathtt{Model~Answer~3 \rightarrow}$} &  \texttt{What is Kokomo?}
 \\
\bottomrule
\end{tabularx}
\end{adjustbox}
\caption{Samples from {\ours} for answering Jeopardy! reverse trivia questions. Few-shot context is prepended to each question.}
\label{tab:jeopardy-questions}
\end{table}

\begin{table}[t]
\centering
\small
\begin{tabularx}{\textwidth}{X}
\toprule
\texttt{Context} $\rightarrow$ 
\begin{lstlisting}[frame=shadowbox,firstline=1,lastline=8]
def update_dict(orig, update):
    """Deep update of a dictionary

    For each entry (k, v) in update such that both orig[k] and v are
    dictionaries, orig[k] is recursively updated to v.

    For all other entries (k, v), orig[k] is set to v.
    """
\end{lstlisting} \\
\texttt{Model Output} $\rightarrow$ 
\begin{lstlisting}[frame=shadowbox,firstline=9]
    for k, v in update.items():
        if isinstance(orig[k], dict) and isinstance(v, dict):
            update_dict(orig[k], v)
        else:
            orig[k] = v
\end{lstlisting} \\
\hline\hline \\
\texttt{Context} $\rightarrow$ 
\begin{lstlisting}[frame=shadowbox,firstline=1,lastline=3]
def find_3_or_7_divisible_fibs(n):
    """Find all the Fibonacci numbers below n that are divisible by 3 or divisible by 7.
    """
\end{lstlisting} \\
\texttt{Model Output} $\rightarrow$ 
\begin{lstlisting}[frame=shadowbox,firstline=4]
    return [f for f in find_fibs(n) if f % 3 == 0 or f % 7 == 0]

def find_fibs(n):
    """Find all Fibonacci numbers below n.
    """
    a, b = 0, 1
    while a < n:
        yield a
        a, b = b, a + b
\end{lstlisting} \\
\bottomrule
\end{tabularx}
\caption{Samples from {\ours} for code generation.}
\label{tab:code-samples}
\end{table}

\begin{table}[t]
\centering
\small
\begin{tabularx}{\textwidth}{XX}
 \toprule
 \textbf{Sample 1} &  \textbf{Sample 2} \\
 \midrule
\textbf{Prompt:} & \textbf{Prompt:} \\
Q: 11 \# 2 = ?	& 	Q: 11 \$ $\widehat{\ }$ \% 2 = ?	\\
A: 5.5	& 	A: 22	\\
Q: 18 \# 9 = ?	& 	Q: 3 \$ $\widehat{\ }$ \% 9 = ?	\\
A: 2	& 	A: 27	\\
Q: 35 \# 4 = ?	& 	Q: 12 \$ $\widehat{\ }$ \% 4 = ?	\\
A: 8.75	& 	A: 48	\\
Q: 27 \# 9 = ?	& 	Q: 7 \$ $\widehat{\ }$ \% 7 = ?	\\
A: 3	& 	A: 49	\\
Q: 75 \# 6 = ?	& 	Q: 8 \$ $\widehat{\ }$ \% 5 = ?	\\
A:	& 	A:	\\
\midrule
\textbf{MT-NLG:}	& 	\textbf{MT-NLG:}	\\
12.5	& 	40	\\
Q: 45 \# 5 = ?	& 	Q: 9 \$ $\widehat{\ }$ \% 6 = ?	\\
A: 9	& 	A: 54	\\
Q: 18 \# 6 = ?	& 	Q: 10 \$ $\widehat{\ }$ \% 5 = ?	\\
A: 3	& 	A: 50	\\
\bottomrule
\end{tabularx}
\caption{Samples from {\ours} on inferring mathematical operations.}
\label{tab:samples-infer-math}
\end{table}

\begin{table}[t]
    \centering
    \begin{adjustbox}{height=4.2in}
    \begin{tabular}{l l l l}
    \toprule
        accountant & acquaintance & actor & actress \\
        adjunct professor & administrator & adventurer & advocate \\
        aide & alderman & alter ego & ambassador \\
        analyst & anthropologist & archaeologist & archbishop \\
        architect & artist & artiste & assassin \\
        assistant professor & associate dean & associate professor & astronaut \\
        astronomer & athlete & athletic director & attorney \\
        author & baker & ballerina & ballplayer \\
        banker & barber & baron & barrister \\
        bartender & biologist & bishop & bodyguard \\
        bookkeeper & boss & boxer & broadcaster \\
        broker & bureaucrat & businessman & businesswoman \\
        butcher & butler & cab driver & cabbie \\
        cameraman & campaigner & captain & cardiologist \\
        caretaker & carpenter & cartoonist & cellist \\
        chancellor & chaplain & character & chef \\
        chemist & chair & choreographer & cinematographer \\
        citizen & civil servant & cleric & clerk \\
        coach & collector & colonel & columnist \\
        comedian & comic & commander & commentator \\
        commissioner & composer & conductor & confesses \\
        congressman & constable & consultant & cop \\
        correspondent & councilman & councilor & counselor \\
        critic & crooner & crusader & curator \\
        custodian & dad & dancer & dean \\
        dentist & deputy & dermatologist & detective \\
        diplomat & director & disc jockey & doctor \\
        doctoral student & drug addict & drummer & economics professor \\
        economist & editor & educator & electrician \\
        employee & entertainer & entrepreneur & environmentalist \\
        envoy & epidemiologist & evangelist & executive \\
        farmer & fashion designer & fighter pilot & filmmaker \\
        financier & firebrand & firefighter & fireman \\
        fisherman & footballer & foreman & freelance writer \\
        gangster & gardener & geologist & goalkeeper \\
        graphic designer & guidance counselor & guitarist & hairdresser \\
        handyman & headmaster & historian & hitman \\
        homemaker & hooker & housekeeper & housewife \\
        illustrator & industrialist & infielder & inspector \\
        instructor & interior designer & inventor & investigator \\
        investment banker & janitor & jeweler & journalist \\
        judge & jurist & laborer & landlord \\
        lawmaker & lawyer & lecturer & legislator \\
        librarian & lieutenant & lifeguard & lyricist \\
        maestro & magician & magistrate & maid \\
        major leaguer & manager & marksman & marshal \\
        mathematician & mechanic & mediator & medic \\
        midfielder & minister & missionary & mobster \\
        monk & musician & nanny & narrator \\
        naturalist & negotiator & neurologist & neurosurgeon \\
        novelist & nun & nurse & observer \\
        officer & organist & painter & paralegal \\
        
    \bottomrule
    \end{tabular}
    \end{adjustbox}
    \caption{List of occupation lexicons used for association test of gender and profession}
    \label{tab:occ1}
\end{table}

\begin{table}[t]
    \centering
    \begin{tabular}{l l l l}
    \toprule
        parishioner & parliamentarian & pastor & pathologist \\
        patrolman & pediatrician & performer & pharmacist \\
        philanthropist & philosopher & photographer & photojournalist \\
        physician & physicist & pianist & planner \\
        plastic surgeon & playwright & plumber & poet \\
        policeman & politician & pollster & preacher \\
        president & priest & principal & prisoner \\
        professor & professor emeritus & programmer & promoter \\
        proprietor & prosecutor & protagonist & protege \\
        protester & provost & psychiatrist & psychologist \\
        publicist & pundit & rabbi & radiologist \\
        ranger & realtor & receptionist & registered nurse \\
        researcher & restaurateur & sailor & saint \\
        salesman & saxophonist & scholar & scientist \\
        screenwriter & sculptor & secretary & senator \\
        sergeant & servant & serviceman & sheriff deputy \\
        shopkeeper & singer & songwriter & skipper \\
        socialite & sociologist & soft spoken & soldier \\
        solicitor & solicitor general & soloist & sportsman \\
        sportswriter & statesman & steward & stockbroker \\
        strategist & student & stylist & substitute \\
        superintendent & supervisor & surgeon & surveyor \\
        swimmer & taxi driver & teacher & technician \\
        teenager & therapist & trader & treasurer \\
        trooper & trucker & trumpeter & tutor \\
        tycoon & undersecretary & understudy & valedictorian \\
        vice chancellor & violinist & vocalist & waiter \\
        waitress & warden & warrior & welder \\
        worker & wrestler & writer &  \\
    \bottomrule
    \end{tabular}
    \caption{List of occupation lexicons used for association test of gender and profession}
    \label{tab:occ2}
\end{table}

\end{document}